\pdfoutput=1

\documentclass[11pt]{article}

\usepackage[final]{acl}

\usepackage{times}
\usepackage{latexsym}

\usepackage[T1]{fontenc}

\usepackage[utf8]{inputenc}

\usepackage{microtype}

\usepackage{inconsolata}

\usepackage{graphicx}
\usepackage{placeins}
\usepackage{float}
\usepackage{tcolorbox}

\usepackage{rotating}
\usepackage{booktabs}
\usepackage{multirow}
\usepackage{tabularx}
\usepackage{threeparttable}
\usepackage{tabu}
\usepackage{enumitem}
\usepackage{subcaption}
\usepackage[table]{xcolor}
\usepackage{makecell}

\title{MAGIC: A Multi-Hop and Graph-Based Benchmark for Inter-Context Conflicts in Retrieval-Augmented Generation}

\author{
Jungyeon Lee\textsuperscript{1 \dag}, Kangmin Lee\textsuperscript{1 2 \dag}, Taeuk Kim\textsuperscript{1 *} \\
\textsuperscript{1}Hanyang University, \textsuperscript{2}KT Corporation \\
{\tt \{jungyune,kimtaeuk\}@hanyang.ac.kr}, {\tt lee.kangmin@kt.com}
}

\begin{document}
\maketitle
\begin{abstract}
Knowledge conflict often arises in retrieval-augmented generation (RAG) systems, where retrieved documents may be inconsistent with one another or contradict the model’s parametric knowledge.
Existing benchmarks for investigating the phenomenon have notable limitations, including a narrow focus on the question answering setup, heavy reliance on entity substitution techniques, and a restricted range of conflict types. 
To address these issues, we propose a knowledge graph (KG)-based framework that generates varied and subtle conflicts between two similar yet distinct contexts, while ensuring interpretability through the explicit relational structure of KGs.
Experimental results on our benchmark, MAGIC, provide intriguing insights into the inner workings of LLMs regarding knowledge conflict: both open-source and proprietary models struggle with conflict detection---especially when multi-hop reasoning is required---and often fail to pinpoint the exact source of contradictions.
Finally, we present in-depth analyses that serve as a foundation for improving LLMs in integrating diverse, sometimes even conflicting, information. 
\end{abstract}

\renewcommand{\thefootnote}{} 
\footnotetext{\textsuperscript{\dag}Equal contribution. \textsuperscript{*}Corresponding author.}
\renewcommand{\thefootnote}{\arabic{footnote}} 




\section{Introduction}

Retrieval-augmented generation (RAG) has become the de facto standard for enhancing the performance of large language models (LLMs) by enabling updates to outdated knowledge and facilitating adaptation to specialized domains \cite{lewis2020retrieval}.
While effective, RAG’s heavy reliance on retrieval quality introduces inherent risks. 
For instance, knowledge obtained from external sources may conflict with the model’s parametric knowledge or contain inconsistencies within the retrieved documents themselves.

\textbf{Knowledge conflict (KC)} is a recent research topic that covers issues related to the aforementioned scenarios and has been receiving attention in the field \cite{xu-etal-2024-knowledge-conflicts}.
An ideal LLM-based system should be robust against knowledge conflict, capable of integrating information from multiple sources—including those that may be contradictory—and ultimately generating reliable responses.
However, its implementation is largely hindered by the challenge of detecting whether disagreements exist across different knowledge sources and, if so, pinpointing exactly where they occur.

\begin{figure}[t]
\centering
\includegraphics[width=0.99\columnwidth]{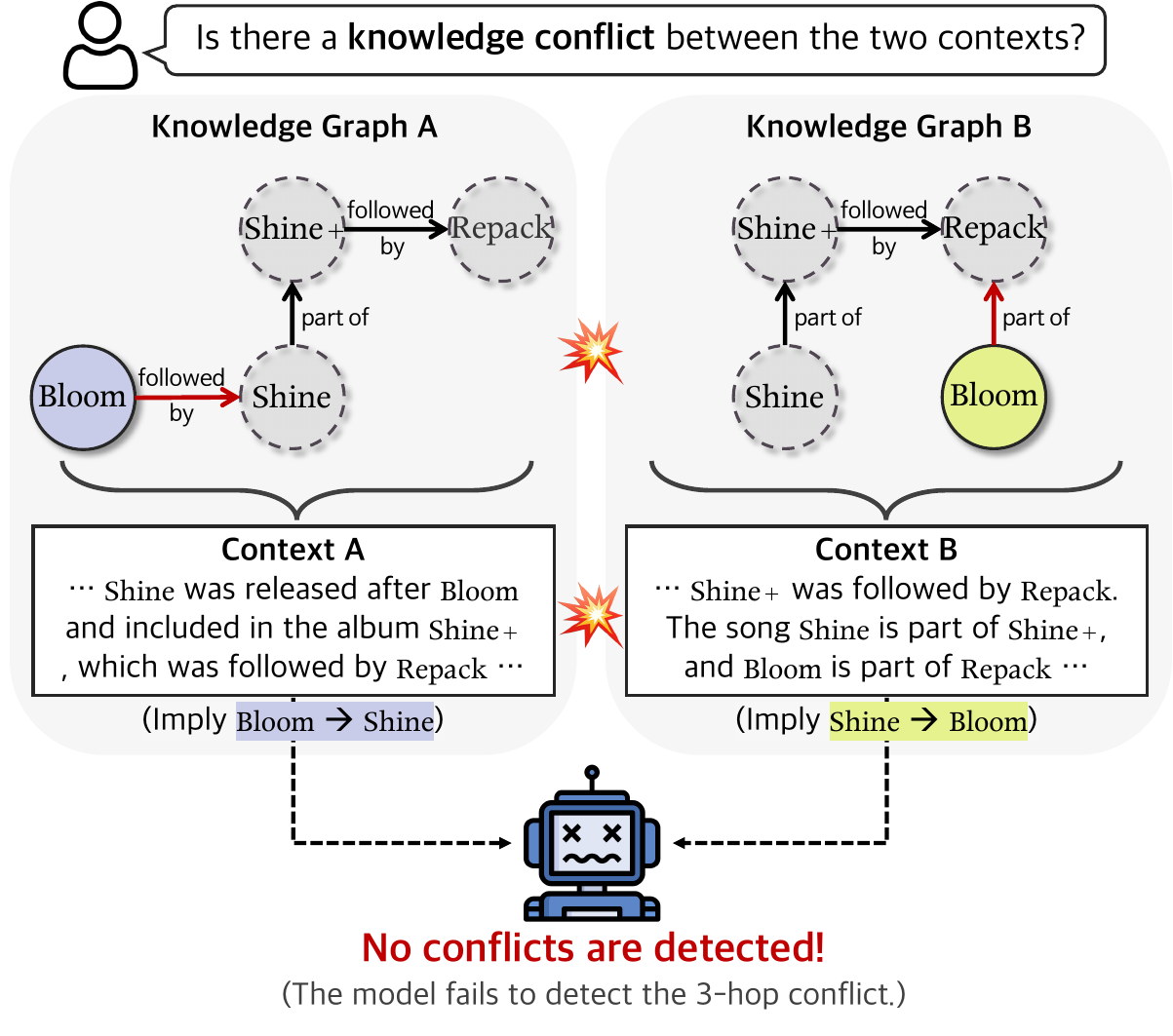}
  \caption{Example of a three-hop conflict from our benchmark, MAGIC. Even advanced LLMs struggle to detect subtle inconsistencies across two contexts, such as conflicting release orders of two songs.}
\label{fig:intro}
\end{figure}

Numerous benchmarks have been introduced to evaluate the performance of LLMs in knowledge conflict detection \cite{9671319, li-etal-2024-contradoc, jiayang-etal-2024-econ, hou2024wikicontradict}. 
Nonetheless, we emphasize that existing research in this area has notable limitations. First, previous studies primarily focus on the question answering (QA) task, where conflicts occur only among multiple answer candidates for a given question \cite{chen-etal-2022-rich, xie2024adaptive, marjanovic-etal-2024-dynamicqa}.
Second, prior research often relies on overly simplistic techniques for dataset construction, e.g., entity substitution  \cite{longpre-etal-2021-entity, chen-etal-2022-rich}, which are insufficient to capture the complex and subtle nature of knowledge conflicts.
Third, while some studies attempt to categorize types of knowledge conflict \cite{hou2024wikicontradict, marjanovic-etal-2024-dynamicqa}, systematic analysis distinguishing between forms, such as single-hop vs. multi-hop conflicts, is still lacking.
Finally, existing benchmarks are largely concerned with exploring conflicts between parametric and external knowledge, while conflicts among multiple input documents remain underexplored \cite{jiayang-etal-2024-econ, hou2024wikicontradict}.\footnote{Knowledge conflict is commonly studied in three setups \cite{xu-etal-2024-knowledge-conflicts}: (1) \textit{context–memory conflict}, involving inconsistencies between parametric and external knowledge; (2) \textit{inter-context conflict}, where contradictions exist among multiple input documents; and (3) \textit{intra-memory conflict}, where inconsistent responses reflect variation in an LLM’s training data. In this work, we focus on \textbf{\textit{inter-context conflict}}.}

To alleviate these issues, we propose a framework for constructing a benchmark targeting \textbf{\textit{inter-context knowledge conflict}}. 
It leverages \textbf{knowledge graphs (KGs)} as the primary source, from which subgraphs are extracted to serve as the basis for distinct knowledge chunks. 
These subgraphs are then perturbed by modifying nodes and edges to introduce conflicts. Finally, both original and altered graphs are converted into corresponding text passages using KG-to-text generation algorithms.

By design, the proposed framework offers several advantages. Leveraging the relational structure of KGs enables greater diversity, complexity, and control in inducing conflicts within documents. 
Moreover, compared to text-based strategies, our approach improves interpretability and supports structured analysis by representing conflicting entities and relations in graph form (see Figure \ref{fig:intro}).

Lastly, we conduct extensive analyses using \textbf{MAGIC} (A \textbf{M}ulti-Hop \textbf{A}nd \textbf{G}raph-Based Benchmark for \textbf{I}nter-\textbf{C}ontext Conflicts), a novel dataset constructed through our framework.\footnote{Available at: \href{https://github.com/HYU-NLP/MAGIC}{https://github.com/HYU-NLP/MAGIC}.}
MAGIC features complex inter-context conflict patterns, including simultaneous and multi-hop cases rarely seen in existing benchmarks. 
Experiments on MAGIC yield key insights into how LLMs perceive knowledge conflicts: (1) most models remain imperfect at detecting conflicts, especially when multi-hop reasoning is required; and (2) even when contradictions are detected, models often fail to localize the exact point of conflict.

\section{Related Work}

\paragraph{Knowledge conflict benchmarks}
Early studies on synthetically inducing sentences with knowledge conflicts (KC) primarily rely on entity substitution \cite{chen-etal-2022-rich}, which is often subsequently polished and paraphrased using LLMs \cite{xie2024adaptive, gokul2025contradiction}. 
Regarding domains and tasks, most prior work focuses on potential inconsistencies arising in QA, such as contradictory answers provided for a given question \cite{longpre-etal-2021-entity}.
However, this paradigm has a clear limitation: most conflicts must appear near answer-related contexts, which strictly limits the range of possible variations.
To overcome this, we adopt question-free, KG-based methods, which inherently support more diverse problem types.\footnote{Concurrently, \citet{gokul2025contradiction} propose a query-free method but rely solely on LLMs, unlike our KG-based one.}

Inter-context knowledge conflict detection evaluates an LLM’s ability to identify contradictions either across multiple input contexts \cite{jiayang-etal-2024-econ, hou2024wikicontradict, marjanovic-etal-2024-dynamicqa} or within a single document \cite{li-etal-2024-contradoc}. 
Existing benchmarks follow two main strategies: collecting conflicts curated from Wikipedia \cite{hou2024wikicontradict} and generating artificial ones using LLMs \cite{jiayang-etal-2024-econ}.
The former offers realism but limited coverage, while the latter is scalable yet often less natural and less representative of real-world scenarios. 
We aim to combine the strengths of both by leveraging the factual grounding of KGs and the generative fluency of LLMs, ensuring both high quality and scalability.
Finally, we highlight that little work has examined performance with respect to fine-grained conflict types. 
Departing from the common convention of using only two conflict categories,\footnote{\citet{jiayang-etal-2024-econ}: answer \& factoid; \citet{hou2024wikicontradict}: explicit \& implicit; \citet{marjanovic-etal-2024-dynamicqa}: static \& dynamic.} we organize our dataset by the number of conflicts and the reasoning hops required for resolution, enabling more systematic analysis.

\paragraph{KG-based dataset creation}
KGs play a crucial role in diverse tasks, e.g., fact verification \cite{kim-etal-2023-kg}, QA \cite{chen2024llm}, and RAG \cite{sanmartin2024kg}, by providing structured representations of knowledge.
In addition, KGs can serve as valuable resources for dataset construction.
For instance, \citet{meng2022locating} introduce COUNTERFACT, a dataset designed to evaluate factual consistency and modifications in LLMs.
In this paper, we also utilize KGs as a foundation for generating realistic and nuanced conflict statements.
\citet{bi2024context} shares similarities with our work, as it also employs Wikidata triplets to induce knowledge conflicts. 
However, their approach is limited to retrieving seed entities for substitution rather than leveraging the full structure of KGs.


\begin{figure}
    \centering
        \includegraphics[width=\linewidth]{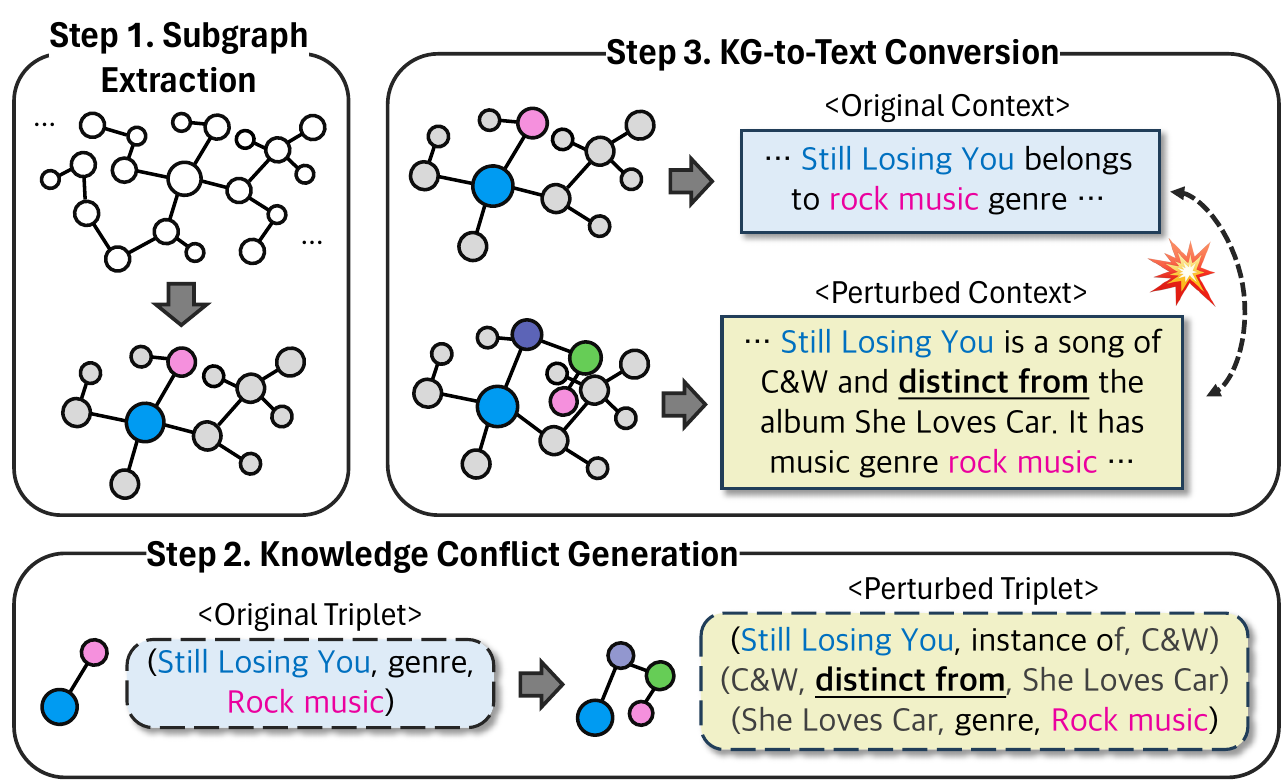} 
        \caption{
        Overview of the proposed KG-based framework for benchmarking inter-context knowledge conflict detection. It comprises three steps: (1) Subgraph Extraction, (2) Knowledge Conflict Generation, and (3) KG-to-Text Conversion, with details listed in \S\ref{sec:MAGIC}.
        }
        \label{fig:overview}
\end{figure}

\section{MAGIC: Multi-Hop and Graph-Based Benchmark for Inter-Context Conflicts} 
\label{sec:MAGIC}

In this section, we introduce a new framework for constructing an inter-context knowledge conflict detection dataset through the collaboration of KGs and LLMs. Compared to existing benchmarks, the proposed approach offers several advantages:
(1) Rather than using QA datasets as the source, we adopt KGs, enabling broader domain coverage and a richer range of conflict forms, including multi-hop cases.
(2) It combines the strengths of both manual and automated strategies by incorporating a human-in-the-loop process during LLM-based generation, ensuring both high quality and scalability.
(3) It also enhances interpretability for users by providing two complementary views, one from the graph and the other from the corresponding text.
 
The procedure consists of three steps, as depicted in Figure \ref{fig:overview}. First, subgraphs are extracted from a KG based on predefined criteria, acting as conceptual knowledge chunks (\S \ref{subsec:subgraph extraction}). Next, perturbations are applied to subgraphs to provoke conflicts (\S \ref{subsec:knowledge conflict generation}). Finally, both original and modified graphs are converted into text passages using KG-to-text algorithms (\S \ref{subsec:kg-to-text}).
As a result, we present a benchmark named \textbf{MAGIC} (A \textbf{M}ulti-Hop \textbf{A}nd \textbf{G}raph-Based Benchmark for \textbf{I}nter-\textbf{C}ontext Conflicts).

\begin{figure}[t]
    \centering
        \includegraphics[width=0.95\linewidth]{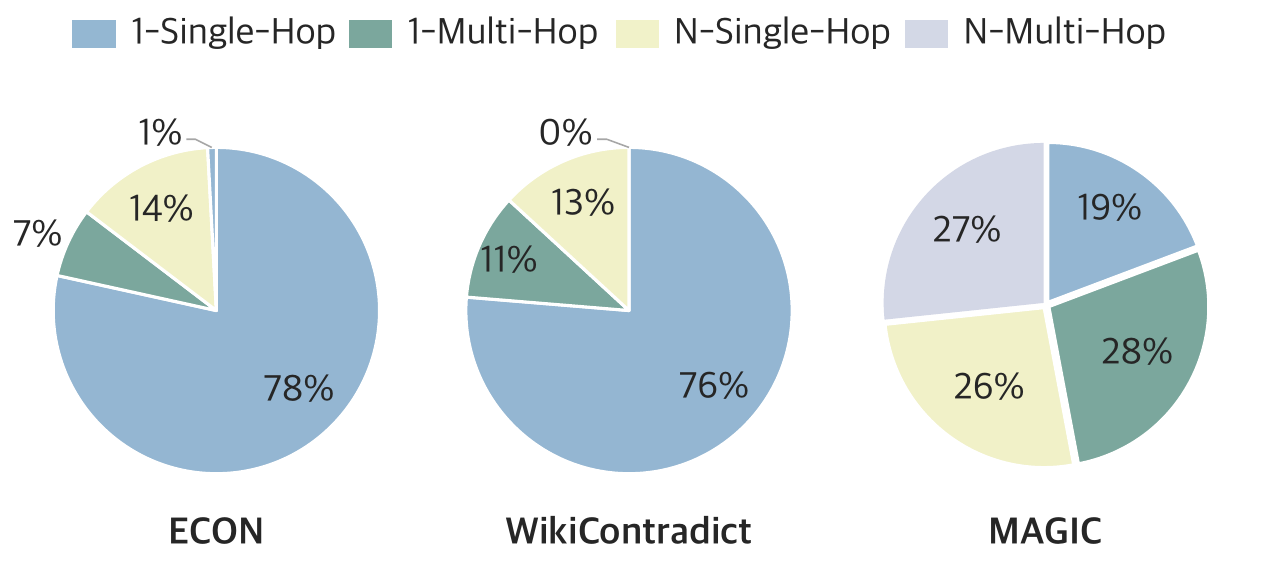} 
        \label{fig:conflict_type_division}
    \caption{Distribution of conflict types across three knowledge conflict detection datasets. MAGIC demonstrates greater diversity and complexity than the others.}
    \label{fig:previous_kc}
\end{figure}

\subsection{Subgraph Extraction}
\label{subsec:subgraph extraction}

As the first step, we distill parts of a large-scale KG to build knowledge segments that serve as targets for inducing knowledge conflicts.
Theoretically, any KG can be utilized; in this work, we employ Wikidata5M \cite{wang-etal-2021-kepler}.
Wikidata5M consists of approximately 20 million triplets, covering various domains and knowledge structures.\footnote{For the diversity and robustness of MAGIC, we preprocess Wikidata5M as follows.
Entities hard to functionally define, e.g., emoticons and special symbols (4,000 in total), are removed.
In addition, general concepts and nodes with too many connections---e.g., ‘human’ and ‘United States’---are excluded.
The 30 most connected nodes are filtered out.}

The key stages of subgraph extraction include seed triplet selection, graph traversal, and enforcing structural constraints.

\paragraph{Seed triplet selection} 
We randomly sample seed triplets that form the basis for subgraph construction. 
Since these triplets define the topic and structure of the resulting subgraphs, we filter the relations they contain. Specifically, from the 825 unique relations in Wikidata5M, we select 46 based on two criteria: (1) semantic clarity, which allows for controlled conflict manipulation, and (2) the ability to support meaningful multi-hop reasoning chains. 
To facilitate more detailed analysis, we group the selected relations into seven semantic domains based on their meaning and typical usage.\footnote{The domains are: Human, Geography, Organization, Creative Work, Class/Concept, Cause-Effect, and General. See Appendix~\ref{sec:relation_list} for the full list of relations used in each domain.}

\paragraph{Graph traversal} Given the seed triplets, we perform graph traversal in the base KG starting from the subject entity of each seed. 
We use the Depth-First Search (DFS) algorithm to progressively expand the subgraph.
DFS is well-suited for exploring deep structural variations within the KG.

\paragraph{Enforcing structural constraints} We regulate DFS traversal with the following constraints to preserve both structural complexity and contextual coherence in the extracted subgraphs.
\begin{itemize}[leftmargin=10pt]
    \item The number of edges in each extracted subgraph is capped at 15 to ensure computational feasibility and maintain interpretability.
    \item  To prevent excessive connectivity, we limit the number of edges per node to 5. This ensures subgraphs retain structural diversity without being dominated by a few highly connected nodes.
    \item The maximum traversal depth of DFS is randomly determined for each run, resulting in subgraphs with diverse diameters and structures.
\end{itemize}

\begin{figure}[t]
  \centering
    \subfloat[1-Single-Hop]{
    \includegraphics[width=0.45\linewidth]{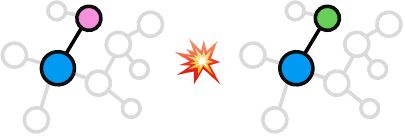}
    }
    \hfill
    \subfloat[N-Single-Hop]{
    \includegraphics[width=0.45\linewidth]{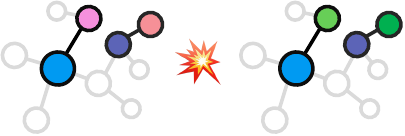}
    }
    \vspace{0.3em} 
    \subfloat[1-Multi-Hop]{
    \includegraphics[width=0.45\linewidth]{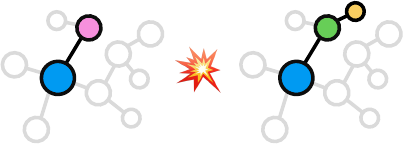}
    }
    \hfill
    \subfloat[N-Multi-Hop]{
    \includegraphics[width=0.45\linewidth]{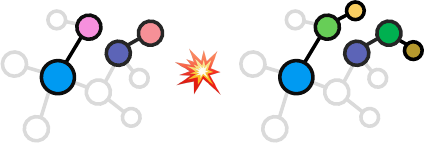}
    }
  \caption{Four distinct types of conflicts in MAGIC.}
  \label{fig:conflict_types}
\end{figure}

\subsection{Knowledge Conflict Generation}
\label{subsec:knowledge conflict generation}

In the second phase, the goal is to perturb and modify extracted subgraphs to create counterparts that contradict the original. 
To this end, we leverage LLMs with strong reasoning capabilities, expecting them to generate plausible and creative candidates that introduce knowledge conflicts within a given context.\footnote{Beyond rule-based substitutions at the entity or relation level as commonly explored in prior work, we aim to induce more natural and contextually coherent conflicts using LLMs. } 
However, na\"ively using such models does not guarantee success, as they are inherently imperfect at recognizing knowledge conflicts.
We therefore propose a method to guide LLMs in reliably generating contradictory facts.

\paragraph{Category of conflicts} 
As illustrated in Figure~\ref{fig:previous_kc}, prior benchmarks, i.e., ECON \cite{jiayang-etal-2024-econ} and WikiContradict \cite{hou2024wikicontradict}, primarily target simple \textit{1-Single-Hop} conflicts (Figure \ref{fig:conflict_types}(a)) arising from individual facts. However, real-world discrepancies often involve multi-hop reasoning and multiple conflicts. To bridge this gap, we define eight distinct conflict types along two axes: (1) the number of reasoning hops required (\textbf{Single-Hop} vs. \textbf{Multi-Hop}) and (2) the number of conflicts present across the two contexts (\textbf{1} vs. \textbf{N}).
\begin{itemize}[leftmargin=10pt]
    \item \textbf{Single-Hop} conflicts (Figure \ref{fig:conflict_types}(a), (b)) arise from inconsistencies through a single relation.
    \item \textbf{Multi-Hop} conflicts (Figure \ref{fig:conflict_types}(c), (d)) require reasoning over multiple connected triplets.
\end{itemize}
Each conflict type is further divided into \textbf{1-conflict} and \textbf{N-conflict} cases, based on the number of contradictions observed between two given contexts, yielding four distinct types illustrated in Figure~\ref{fig:conflict_types}. 
The number of data instances allocated to each category is shown in Table \ref{tab:data_statistics}. 
This categorization enables fine-grained evaluation across varying reasoning depths and conflict complexities.

\begin{table}
    \centering
    \footnotesize
    \setlength{\tabcolsep}{0.3em}
    \begin{tabular}{l c c c c c c c c c}
    \toprule
        \textbf{Conflict Type}
        & \multicolumn{4}{c}{\textbf{Single-Hop}} 
        & \multicolumn{4}{c}{\textbf{Multi-Hop}} 
        & \multirow{2}{*}{\shortstack[c]{\textbf{Total}}} \\
        \cmidrule(lr){2-5} \cmidrule(lr){6-9}
        \textbf{\# Conflict} & \textbf{1} & \textbf{2} & \textbf{3} & \textbf{4} & \textbf{1} & \textbf{2} & \textbf{3} & \textbf{4}\\
    \midrule
        \textbf{\# Instances} & 208 & 154 & 80 & 50 & 300 & 158 & 80 & 50 & 1,080 \\
    \bottomrule
    \end{tabular}
    \caption{MAGIC dataset statistics by conflict category.}
    \label{tab:data_statistics}
\end{table}

\paragraph{Subgraph-level few-shot prompting}
\label{para:subgraph few shot}
We use OpenAI’s o3-mini \cite{o3-mini} to induce and collect conflict candidates.
Given a target seed triplet, we prompt the LLM with both the triplet and its surrounding context, represented as a set of subject–relation–object triplets from the subgraph. 
Including the surrounding subgraph allows the model to generate natural contradictions that remain consistent with the local context.\footnote{Appendix~\ref{sec:triplet_subgraph} reveals that using only the target triplet often produced contextually misaligned conflicts.}

Still, we observe that prompting without task demonstrations (i.e., real examples of knowledge conflicts) often falls short in generating diverse and logically complex conflicts, particularly in multi-hop scenarios.\footnote{The model tends to repeat patterns, struggling to produce varied conflicts and often misaligning with common sense.} As a solution, we adopt a few-shot prompting strategy, where the prompt includes three validated conflict examples per relation type. This encourages the model to move beyond simple entity or relation swaps. The final prompt template used in this process is shown in Figure~\ref{fig:prompt_kg_generation}.\footnote{While designed for multi-hop conflicts, the template can be readily adapted to single-hop settings.} 
For N-conflict cases, the same graph is reused with different perturbations until the desired number of conflicts is achieved.

\begin{figure}[h]
    \setlength{\fboxsep}{7pt} 
    \noindent
    \begin{tcolorbox}[
        colback=pink!10,    
        colframe=pink!75,   
        sharp corners=south, 
        rounded corners=northwest,
        boxrule=0.8pt,      
        width=\columnwidth, 
        fonttitle=\bfseries,
        coltitle=black,     
        title=Knowledge Conflict Generation Prompt
    ]
        \fontsize{9pt}{11pt}\selectfont 
        \setlength{\parskip}{5pt} 

        \textbf{Instruction} 

        You will be provided with an [Original Triplet] and a set of [Related Subgraphs]. Your task is to generate a \textbf{multi-hop knowledge conflict} consisting of 2–3 logically connected triplets that together create a logically coherent but indirect conflict with the [Original Triplet].

        \textsc{\#\# Requirements:} \\
        - Construct a conflict that does not directly contradict the original triplet, but introduces contradiction through intermediate reasoning steps. \\\vspace{-15pt}

        - Use \textbf{one or more specific entities or relations} from the [Related Subgraphs] to build the multi-hop chain. \\\vspace{-15pt}

        - Each triplet must be semantically valid and form a realistic knowledge path. \\\vspace{-15pt}
        
        - The conflict must be concrete, not vague or overly inferential.

        \textsc{\#\# Output Format:} \\
        Return a set of 2–3 triplets in the form (Subject | Relation | Object) that together form the multi-hop conflict.
        Do not include explanations, reasoning steps, or any additional text. \\\vspace{-5pt}
        
        \textbf{Demonstrations}
        
        \textsc{[Original Triplet]}
    (tocantins (state) |  divides into | novo jardim) \\\vspace{-15pt} 
    
        \textsc{[Modified Triplet]}
    (tocantins (state) | borders | mato grosso) (mato grosso | contains | novo jardim) \\\vspace{-5pt}
        ... \\\vspace{-5pt}
        
    \end{tcolorbox}
    \caption{Prompt for generating multi-hop conflicts.}
    \label{fig:prompt_kg_generation}
\end{figure}

\begin{figure}
    \centering
        \includegraphics[width=\linewidth]{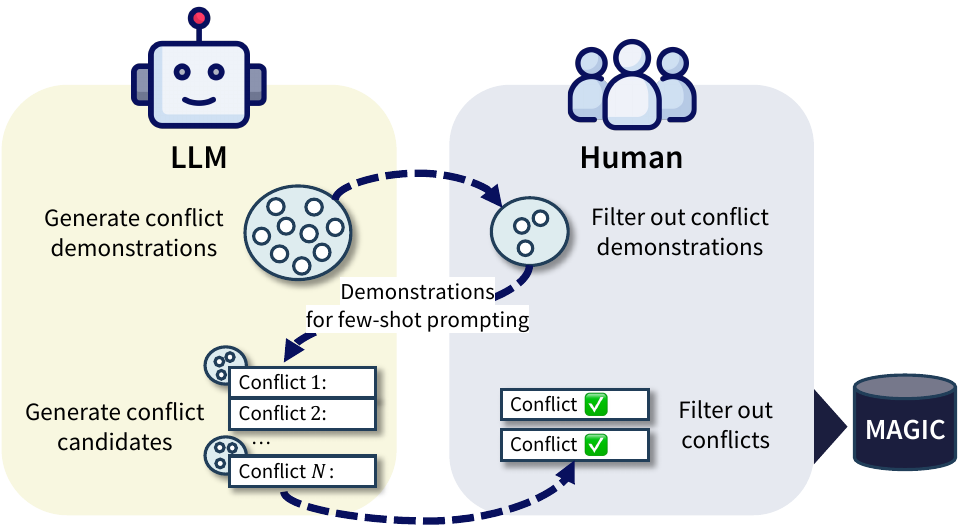} 
        \caption{
        Two-stage human-in-the-loop pipeline for data quality control.
        }
        \label{fig:human-in-the-loop}
\end{figure}

\paragraph{Quality control via human feedback}
Since LLMs are not inherently good at detecting knowledge conflicts, it may seem paradoxical to ask them to generate such cases. 
To alleviate this,  we incorporate a human-in-the-loop process at two stages---before and after few-shot prompting---to improve dataset quality.
First, human experts intervene during few-shot demonstration selection, manually filtering model-generated cases.
Second, experts filter out trivial or incoherent outputs after conflict generation.\footnote{Two researchers independently reviewed disjoint subsets using a shared guideline. See Appendix~\ref{sec:annotation_guideline} for full details.} 
Figure \ref{fig:human-in-the-loop} depicts the entire workflow.

\subsection{KG-to-Text Conversion}
\label{subsec:kg-to-text}

To represent knowledge conflicts from graphs in natural language, we apply KG-to-text conversion, following the approach of \citet{kasner-dusek-2024-beyond} with modifications. 
Using the prompts shown in Figure~\ref{fig:prompt_kg_to_text}, GPT-4o-mini \cite{gpt4o-mini}\footnote{Specific version: gpt-4o-mini-2024-07-18.} generates coherent textual contexts while preserving the semantics of the original graph. 
To ensure transformation accuracy, we perform automatic verification using Claude 3.7 Sonnet \cite{3_7Sonnet}, with the prompt shown in Figure \ref{fig:prompt_kg_to_text_eval}.

In addition, we validate data integrity by sampling and manually inspecting a subset of generated instances.
Our manual inspection confirms that the conversion process achieves consistently high quality; further details are provided in Appendix~\ref{subsec:kg_to_text_prompt}.

\section{Experimental Setups}
\label{sec:experimental setups}
With the MAGIC benchmark as a foundation, we conduct experiments to examine how LLMs handle inter-context knowledge conflicts.
We evaluate various open-source and proprietary LLMs without task-specific training, instead prompting them to identify potential contradictions.
In the following, we outline the LLMs, datasets, prompting strategies, and metrics used in our experiments.

\paragraph{LLMs} We use 5 LLMs: Llama 3.1 70B Instruct \cite{dubey2024llama}, o1 \cite{o1}, Mixtral-8x7B Instruct \cite{mixtral}, Claude 3.5 Haiku \cite{Haiku}, GPT-4o-mini \cite{gpt4o-mini}.

\paragraph{Datasets} 
Alongside MAGIC, we employ two existing benchmarks to highlight its strengths.

\begin{itemize}[leftmargin=10pt]
    \item \textbf{ECON} \cite{jiayang-etal-2024-econ}:  A dataset created by introducing evidence conflicts through two methods—answer conflicts and factoid conflicts—highlighting contradictions in supporting evidence.
    It contains 168 data instances.
    \item \textbf{WikiContradict} \cite{hou2024wikicontradict}: A human-annotated QA benchmark utilizing Wikipedia’s contradiction tags to capture real-world knowledge conflicts. It categorizes contradictions into explicit and implicit types. After deduplication, it comprises 103 data samples.
    \item \textbf{MAGIC}: It is constructed atop KGs, where conflicts are systematically induced from the underlying relational structure. 
    It encompasses both single-hop and multi-hop contradictions, with the number of conflicts dynamically varying across context pairs. 
    In total, MAGIC consists of 1,080 carefully curated examples, with comprehensive statistics presented in Table~\ref{tab:data_statistics}. 
    Remarkably, the scale of MAGIC surpasses that of existing benchmarks by a significant margin, offering a richer and more challenging resource for evaluating inter-context conflict detection.
\end{itemize}

\paragraph{Prompting strategy}
\label{para:prompting strategy}
Prior work \cite{jiayang-etal-2024-econ,hou2024wikicontradict} typically frames the task as a binary classification problem, relying on minimal prompts (see Appendix~\ref{subsec:binary_prompt}). In contrast, we introduce a stepwise prompting strategy (refer to Figure~\ref{fig:multi_step_prompt} for details) designed to more fully probe the capabilities of LLMs for inter-context conflict detection.

\paragraph{Metrics} 
To account for the stochasticity of LLMs, all models perform three independent inference runs. 
We use two metrics for fine-grained evaluation, with scores averaged across all instances in each dataset. 
These metrics are manually computed by participating researchers, as automatic methods---such as LLM-as-a-judge---remain insufficiently reliable for this task.\footnote{If they were, further investigation into knowledge conflict detection would be unnecessary. See Appendix~\ref{sec:human eval} for details.}
\begin{itemize}[leftmargin=10pt]
    \item Identification (\textbf{ID}) score: If a model fails to detect a conflict in any of the three attempts, it receives a score of 0; otherwise, it receives 1.
    \item Localization (\textbf{LOC}) score: We further evaluate LLMs' performance on conflict localization. A full score (1) is awarded only if all conflicting locations are correctly identified; otherwise, the score is 0.
    Note that this fine-grained evaluation has not been considered in previous work.
\end{itemize}

\begin{table}
    \centering
    \small
    \setlength{\tabcolsep}{0.4em}
        \begin{tabular}{l c c >{\columncolor{gray!10}}c}
    \toprule
        \textbf{Models / Datasets} & \textbf{ECON} & \textbf{WikiContradict} & \textbf{MAGIC} \\
    \midrule
        Mixtral 8x7B & 46.43 & 52.43 & \textbf{37.92} \\
        Llama 3.1 70B & 81.41 & 78.79 & \textbf{72.86} \\
        Claude 3.5 Haiku & 83.33 & 61.17 & \textbf{60.28} \\
        GPT-4o-mini & 88.10 & 82.52 & \textbf{83.61} \\
        o1 & 74.40 & 74.76 & \textbf{68.06} \\
    \midrule
        Average & 74.73 & 69.93 & \textbf{64.54} \\
    \bottomrule
    \end{tabular}
    \caption{\textbf{ID} scores (\%) on three KC detection datasets. Lower scores indicate higher task complexity. The lowest value in each row is shown in \textbf{bold}.}
    \label{tab:main_table_ID}
\end{table}

\section{Experimental Results} 
\label{sec:main results}

The main experimental results are shown in Table~\ref{tab:main_table_ID} and Table~\ref{tab:main_table_LOC}. 
Lower scores indicate greater difficulty, suggesting the dataset is more challenging.

\paragraph{Overall results} LLMs tested on MAGIC consistently show lower ID and LOC scores compared to those on ECON and WikiContradict, with average scores decreasing by up to 10\% and 17\%, respectively. 
This indicates that models struggle more to identify conflicts in our dataset, and even when they do, they have difficulty pinpointing the exact portions where the conflict occurs.

\paragraph{ID scores per LLM} 

Table~\ref{tab:main_table_ID} shows that GPT-4o-mini achieves the highest accuracy and generalizes well to MAGIC. 
Mixtral performs the worst overall, while Haiku still lags behind larger models, suggesting difficulty with multi-hop reasoning. 
Llama shows a unique trend on MAGIC: it fails to respond in 47.4\% of runs but achieves high ID scores when it does.
Finally, o1 exhibits moderate performance, implying that conflict detection may depend on more than just LLMs' reasoning ability.

\begin{table}
    \centering
    \small
    \setlength{\tabcolsep}{0.4em}
    \begin{tabular}{l c c >{\columncolor{gray!10}}c}
    \toprule
        \textbf{Models / Datasets} & \textbf{ECON} & \textbf{WikiContradict} & \textbf{MAGIC} \\
    \midrule
        Mixtral 8x7B & 35.71 & 40.78 & \textbf{17.40} \\
        Llama 3.1 70B & 54.49 & 51.52 & \textbf{37.92} \\
        Claude 3.5 Haiku & 66.07 & 52.43 & \textbf{42.50} \\
        GPT-4o-mini & 63.69 & 68.93 & \textbf{55.00} \\
        o1 & 64.88 & 65.48 & \textbf{49.72} \\
    \midrule
        Average & 57.09 & 55.74 & \textbf{40.51} \\
    \bottomrule
    \end{tabular}
    \caption{\textbf{LOC} scores (\%) on three KC detection datasets. Lower scores indicate higher complexity. We observe that models struggle more with pinpointing the exact location of a conflict than with detecting its presence. The lowest value in each row is shown in \textbf{bold}.}
    \label{tab:main_table_LOC}
\end{table}

\begin{figure}
    \centering
        \includegraphics[width=\linewidth]{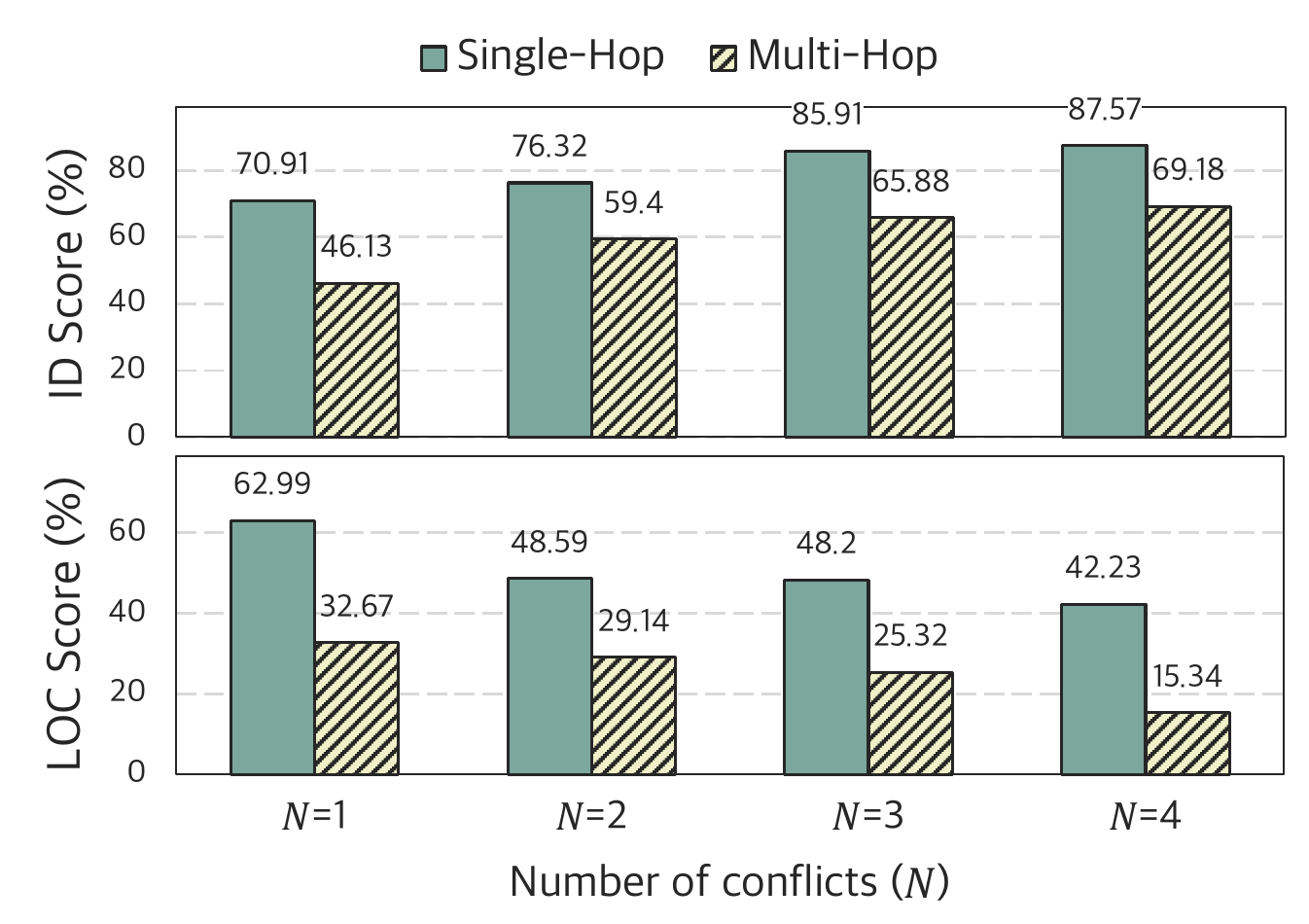} 
        \caption{
        Average LLM performance on MAGIC by conflict type. A greater number of conflicts aids recognition but hinders localization, while multi-hop cases remain inherently more challenging than single-hop.
        }
        \label{fig:data_category}
\end{figure}

\paragraph{LOC scores per LLM} 

Table~\ref{tab:main_table_LOC} shows trends similar to Table~\ref{tab:main_table_ID}, with GPT-4o-mini consistently achieving the best LOC scores across all datasets.
Qualitative analysis reveals that o1 often takes a conservative stance, frequently predicting \textit{no conflict} in ambiguous cases,\footnote{e.g., (A consists of B) vs. (A has C), (B not part of C).} which leads to missed conflicts—particularly those requiring multi-hop thinking (see Table~\ref{tab:detailed_LOC}).\footnote{This outcome is unexpected, given o1’s strong reasoning abilities in math and coding. One possible explanation is the exclusion of explicit reasoning instructions such as ``think step by step''. We leave further investigation on this as future work.
} Mixtral’s low LOC score largely stems from poor initial conflict identification, reducing the number of evaluated localization cases. Llama shows moderate performance but produces overly long outputs (1121.5 tokens vs. 631.1 for others), often including irrelevant content. This suggests over-reliance on instructions, leading to over-detection and reduced localization precision.

\paragraph{Performance by conflict types}

Figure \ref{fig:data_category} presents the average performance of all LLMs across four conflict complexity settings: single-hop vs. multi-hop and 1-conflict vs. N-conflict, as in Section~\ref{subsec:knowledge conflict generation}. 
Detailed results are provided in Appendix~\ref{subsec:detailed_results}. 

Single-hop conflicts, often involving entity or relation substitutions, are relatively easier, with models performing well in both identification and localization. 
In contrast, multi-hop conflicts introduce greater complexity, as contradictions become more indirect, leading to lower ID and LOC scores. 
Localization is especially difficult in multi-hop cases, as conflicts often span multiple locations.

Meanwhile, a higher number of conflicts reflects a stronger contradiction between the two contexts, making conflict detection relatively easier for models. 
In other words, while more conflicts facilitate identification, they also complicate precise localization.
When multiple conflicts occur, pinpointing all the individual conflicting sentences becomes more difficult, thereby leading to lower LOC scores.
A similar trend has also been observed in ECON in Appendix~\ref{sec:econ_multiple} supports this finding.

\begin{figure}[t]
\centering
\includegraphics[width=\linewidth]{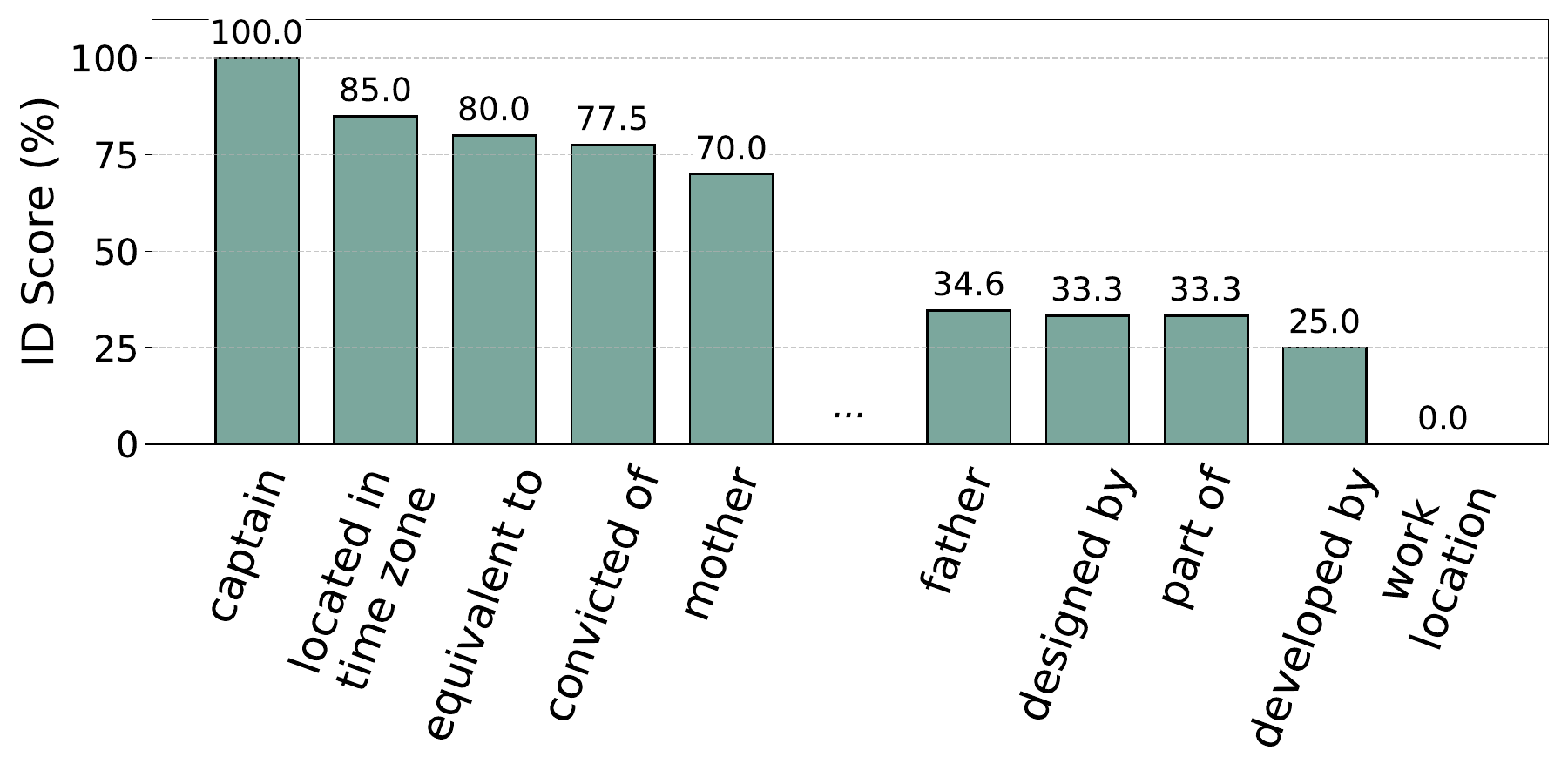}
  \caption{Average performance of LLMs on the 5 most predictive and 5 least predictive relations. This analysis focuses solely on the single-conflict subset of MAGIC.}
\label{fig:perform_by_relation}
\end{figure}

\paragraph{Domain- \& relation-level analysis}
As MAGIC spans diverse topics and relations in KGs, it enables analysis at both the domain and relation levels.
Detailed domain-level statistics are reported in Table~\ref{tab:perform_by_domain}.
In the domain-specific evaluation, we find that most LLMs perform relatively well on the Class/Concept domain, likely because it contains clear and well-defined relations (e.g., \textit{subclass of}, \textit{different from}) that are readily recognizable.\footnote{Still, exceptions exist depending on the used model.}
In contrast, performance on the Organization domain varies widely---from around 75\% (GPT-4o-mini, Llama) to below 45\% (Mixtral, Haiku, o1)---indicating that models differs in handling hierarchical relations.

Meanwhile, Figure~\ref{fig:perform_by_relation} highlights the top five most and least predictive relations across all LLMs. 
Conflicts involving \textit{captain} and \textit{mother} are relatively easy for models, while \textit{work location} and \textit{father} pose greater challenges. 
More representative examples are provided in Table~\ref{tab:easy_hard_example} in the Appendix.

Finally, Figure \ref{fig:perform_by_number_of_domain} shows that detection performance is influenced by the number of domains present in the context.
In multi-domain cases, ID scores tend to improve—likely due to increased semantic diversity making conflicts more salient. 
In contrast, LOC scores decline, possibly because structural variation complicates span localization.

\section{Further Analysis and Discussion}
\label{sec:discussion}

\begin{figure}[t]
\centering
\includegraphics[width=\columnwidth]{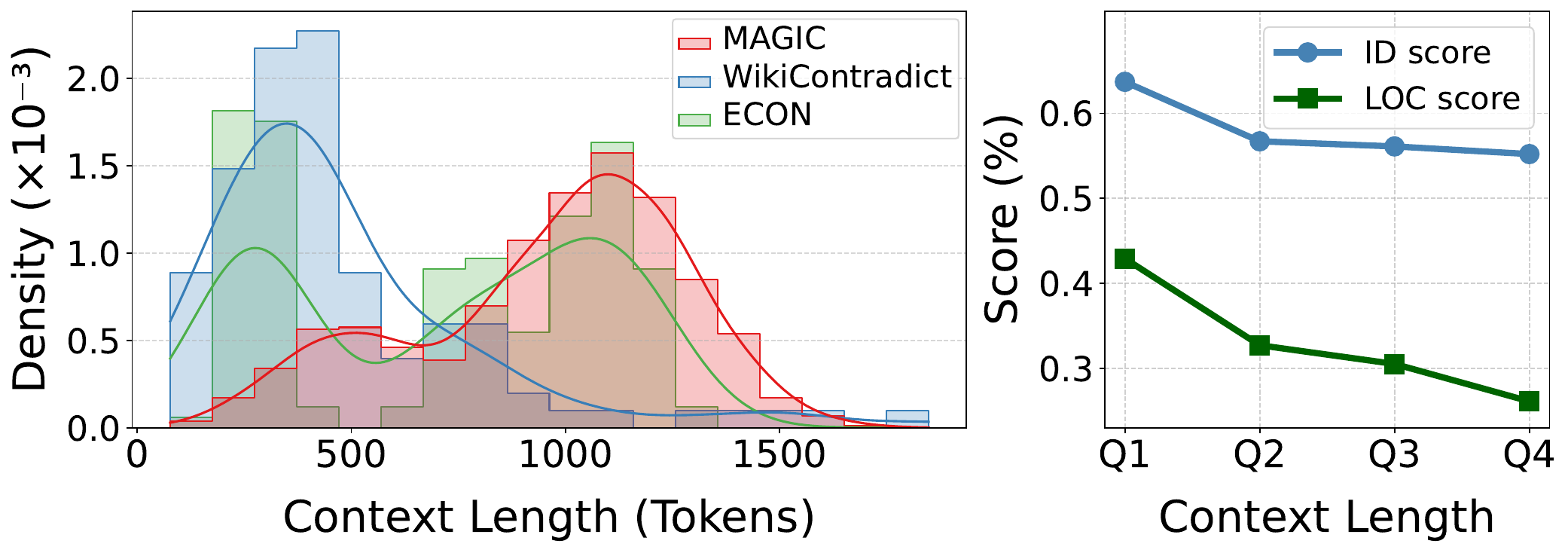}
  \caption{(Left) Context length distributions of the three KC datasets. (Right) ID and LOC scores averaged over LLMs decline as context length increases in MAGIC.}
\label{fig:perform_by_length}
\end{figure}

\paragraph{Taxonomy-based analysis} 
\label{subsec:existing_kc}
To systematically analyze conflict patterns, we apply our proposed conflict typology to ECON and WikiContradict. 
Representing these datasets as graphs allows us to highlight their characteristics.
A key challenge is the lack of predefined ontologies or domain structures, which hinders the use of traditional ontology-based methods \cite{cauter-yakovets-2024-ontology}. To address this, we use LangChain \cite{Langchain} to construct reliable, schema-free KGs that support structured conflict representation.

Figure \ref{fig:previous_kc} shows that 1-Single-Hop conflicts—the typically easiest case—are the most prevalent in prior datasets, accounting for 78\% in ECON and 76\% in WikiContradict.
In contrast, MAGIC exhibits a balanced distribution across conflict types, with substantial proportions of 1-Multi-Hop (28\%) and N-Multi-Hop (27\%), underscoring its robustness as a benchmark.

\paragraph{Length-based analysis}
In conflict detection, a reasonable hypothesis is that context length positively correlates with dataset difficulty, as longer documents tend to involve more complex linguistic structures.
To validate this, we present length-oriented anlysis in Figure \ref{fig:perform_by_length}.
The left panel visualizes the total context length distribution among the three datasets, with MAGIC containing more long-context examples.
We suspect MAGIC’s multi-hop and description-rich design contributes to this.

Further, we group MAGIC into four bins based on context length and report their respective performance in the right panel of Figure~\ref{fig:perform_by_length}.
To balance group sizes, we use quantile-based binning: Q1 contains the shortest contexts, and Q4 the longest.
As context length increases, both scores decline, with LOC dropping more sharply---indicating the growing difficulty of pinpointing conflicting spans in longer inputs.

\begin{figure}[t]
    \includegraphics[width=0.9\linewidth]{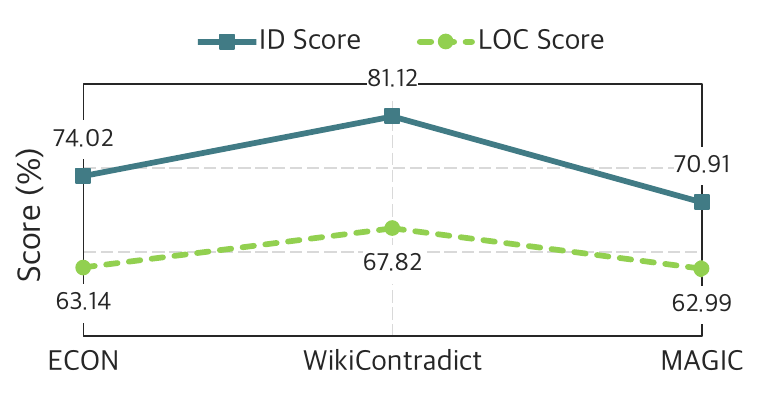}
  \caption{Comparison of ID and LOC scores on 1-Single-Hop conflicts across three datasets.}
  \label{fig:compare_single_hop}
\end{figure}

\paragraph{Comparison with existing KC datasets}
In addition to the results discussed in \S\ref{sec:main results}, we compare MAGIC with existing datasets by focusing on the 1-Single-Hop type—the only category shared across all datasets—to ensure a fair and consistent comparison. 
In ECON and WikiContradict, conflict categories are manually annotated.
Figure \ref{fig:compare_single_hop} shows that even on relatively simple 1-Single-Hop conflicts, models perform worse on MAGIC than on ECON and WikiContradict, suggesting that our dataset poses challenges beyond surface-level contradiction.

\begin{figure}[t]
\centering
\includegraphics[width=0.9\columnwidth]{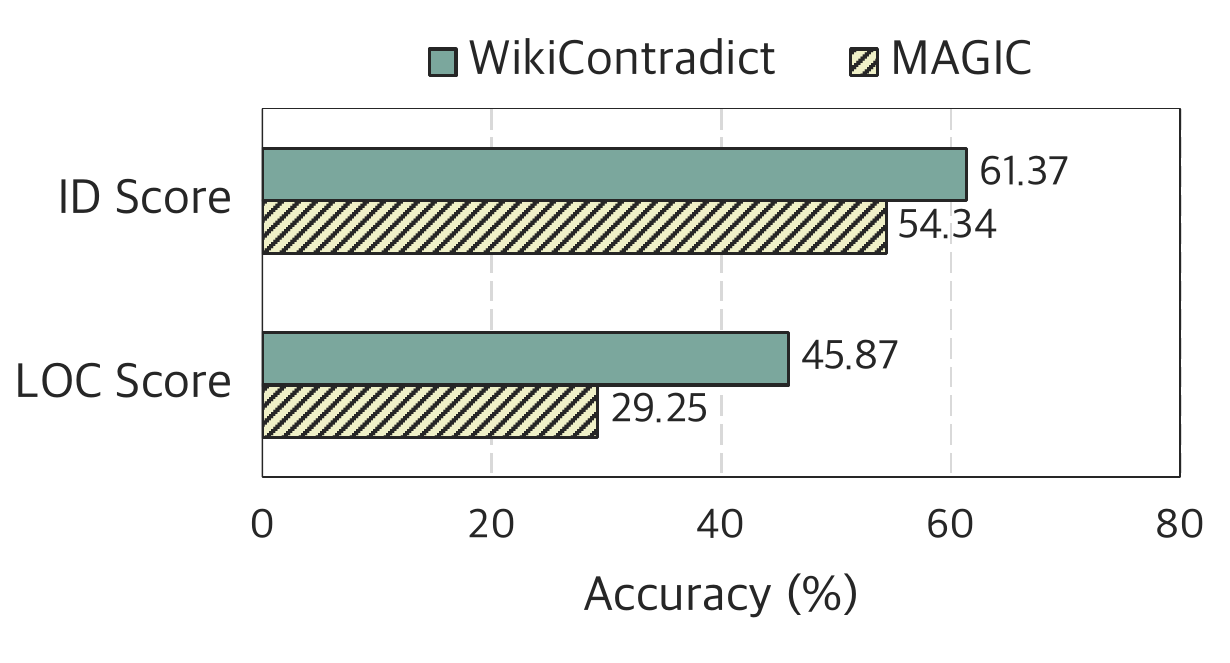}
\caption{Comparison of difficulty between challenging subsets of WikiContradict and MAGIC, with scores averaged over all five LLMs.}
\label{fig:compare_previous}
\end{figure}

We further conduct a comparative study between the challenging subset of WikiContradict (Implicit) and the corresponding subset from our dataset (Multi-Hop in MAGIC).\footnote{WikiContradict consists of explicit and implicit conflicts; the latter are considered more challenging due to their subtlety.}
Figure \ref{fig:compare_previous} reports that MAGIC proves more difficult for LLMs than WikiContradict.
MAGIC yields an ID score up to 9\% lower and a LOC score up to 16\% lower compared to WikiContradict.
This highlights MAGIC’s intrinsic complexity, setting it apart from existing datasets.

\begin{figure}[t]  
\centering
\includegraphics[width=\columnwidth]{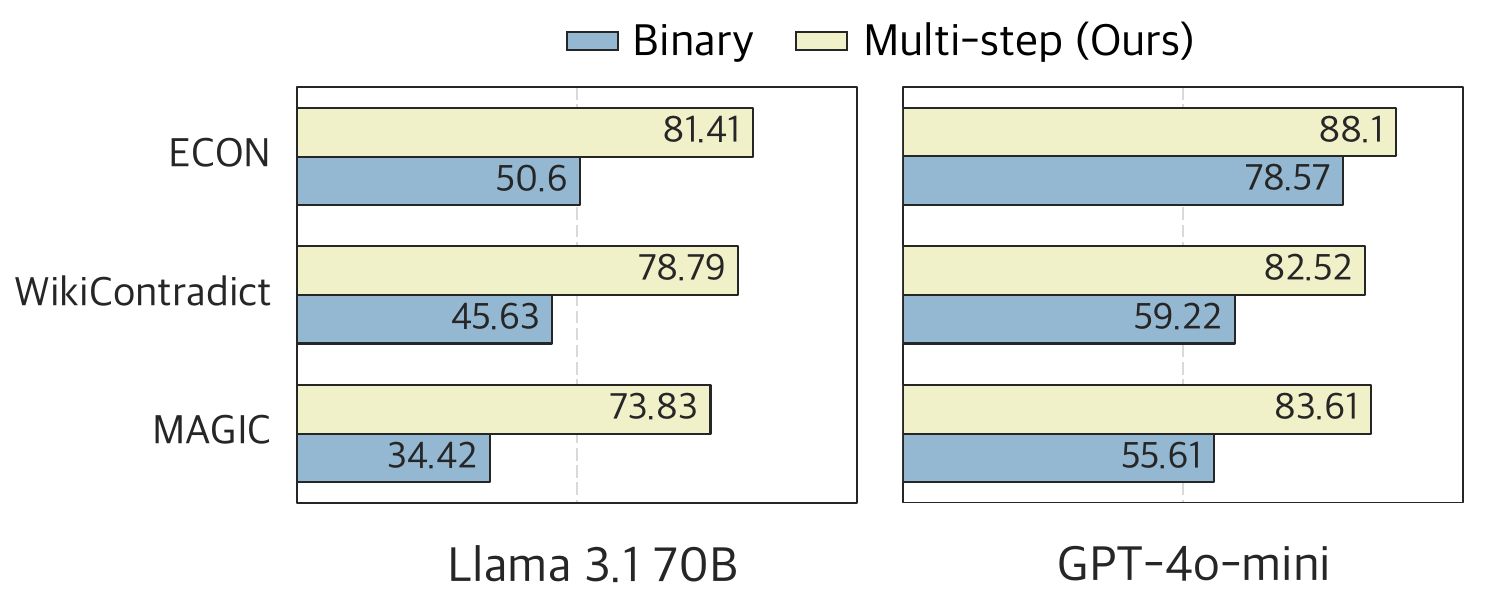}
    \caption{Comparison of ID scores for binary and multi-step prompts, tested on two models.}
\label{fig:compare_detection_prompts}
\end{figure}

\begin{figure}[t]
\centering
\includegraphics[width=0.8\columnwidth]{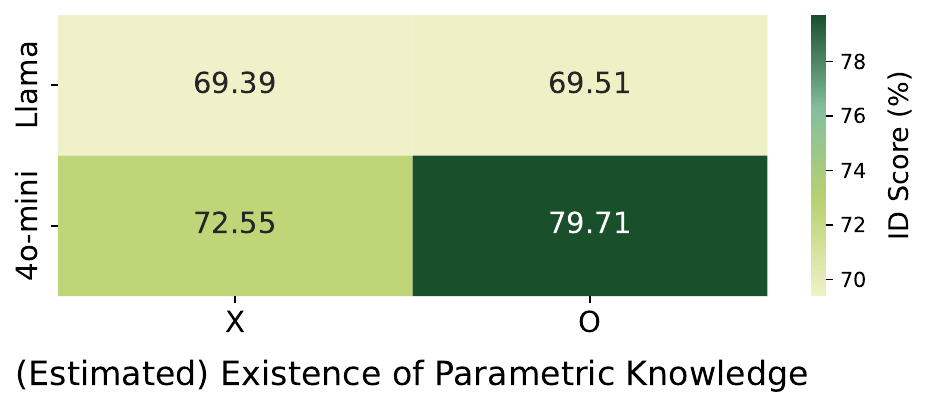}
  \caption{ID scores of two LLMs on known (O) vs. unknown (X) instances, based on their parametric knowledge. Models show slightly stronger conflict detection when the relevant knowledge is already embedded.}
\label{fig:eval_prior_knowledge}
\end{figure}

\paragraph{Impact of prompts} 
While prior work \cite{jiayang-etal-2024-econ,hou2024wikicontradict} typically employs binary (yes/no) prompts for conflict detection—often oversimplifying the task—we use a multi-step technique that helps improve performance.
To compare the effectiveness of these two distinct prompting strategies for conflict detection, we conduct experiments, with results reported in Figure~\ref{fig:compare_detection_prompts}. 
Across all cases, the results show that our multi-step prompting approach outperforms the na\"ive prompt, achieving improvement up to 39.41\%.
This implies that although we employ a method superior to those commonly used in the literature, there remains substantial room for addressing the challenges posed by MAGIC.

\paragraph{Impact of parametric knowledge}
In the literature on \textit{inter-context} conflict, research has primarily focused on conflicts between two input documents, overlooking the influence of parametric knowledge which can significantly affect the performance of LLMs in knowledge conflict detection.

To address this, we explore an underexamined setup by splitting a subset of MAGIC---instances with 1- and 2-conflicts---into two groups based on the estimated presence of parametric knowledge in LLMs.
Concretely, we approximate the existence of parametric knowledge for a given triplet by posing a converted verification question—for example, ‘Is Barack Obama Sr. the father of Barack Obama?’.
We label a triplet as known if the model provides the correct answer in at least 4 out of 5 attempts, and as unknown if it succeeds in no more than 1.

As shown in Figure~\ref{fig:eval_prior_knowledge}, both GPT-4o-mini and Llama 3.1 achieve slightly higher ID scores on instances with prior knowledge.
These findings suggest that models may detect conflicts more effectively when they already possess the relevant factual knowledge, although a more systematic investigation is left for future work.




\paragraph{Impact of context order}
To probe whether context order influences model behavior, we evaluate both the original MAGIC and a variant in which the two contexts in each instance are reversed.

As reported in Table~\ref{tab:context_swap}, the ID scores remain largely consistent across models, with Mixtral 8x7B showing a slight increase and others nearly unchanged.  
These results indicate that context order does not significantly affect model performance on MAGIC, suggesting that our findings are robust to input ordering.

\begin{table}[t]
    \centering
    \small
    \begin{tabular}{l c c >{\columncolor{gray!10}}c}
    \toprule
        \textbf{Models} & \textbf{Original} & \textbf{Swap} & \textbf{$\Delta$ (Swap–Orig)} \\
    \midrule
        Mixtral 8x7B & 37.92 & 46.45 & \textcolor{red}{+8.53} \\
        Llama 3.1 70B & 73.83 & 75.64 & \textcolor{red}{+1.81} \\
        Claude 3.5 Haiku & 61.11 & 61.01 & \textcolor{blue}{-0.10} \\
        GPT-4o-mini & 83.61 & 81.86 & \textcolor{blue}{-1.75} \\
    \bottomrule
    \end{tabular}
    \caption{Effect of context order on ID scores in MAGIC.}
    \label{tab:context_swap}
\end{table}

\paragraph{Linguistic quality of MAGIC}
\label{para:linguistic_quality}
An essential requirement for MAGIC is that its conflict data be both realistic and stylistically similar to human-written text.
To validate this, we use an LLM judge (Claude 3.7 Sonnet) to evaluate two aspects: (i) \textit{naturalness}---whether the text is fluent and human-like---and (ii) \textit{realism}---whether the scenario appears plausible in real-world contexts.
We randomly sample 155 contexts from three datasets to match overall length distributions. 
Each context is rated on a 0–5 scale, with the detailed evaluation prompts provided in Figures \ref{fig:eval_naturalness} and \ref{fig:eval_realism}.

As shown in Table \ref{tab:natural_realism}, MAGIC achieves naturalness scores close to WikiContradict (-0.31), despite WikiContradict relying on real Wikipedia text. 
It further surpasses ECON in realism, showing that MAGIC generates scenarios that are both plausible and human-like. 
These results underscore MAGIC’s reliability as a benchmark for advancing research in knowledge conflict detection.

\begin{table}[t]
    \centering
    \small
    \begin{tabular}{l c c >{\columncolor{gray!10}}c}
    \toprule
        \textbf{Metric} & \textbf{ECON} & \textbf{WikiContradict} & \textbf{MAGIC} \\
    \midrule
        Naturalness & 4.36 & 4.39 & 4.08 \\
        Realism & 4.00 & 4.72 & 4.26 \\
    \bottomrule
    \end{tabular}
    \caption{Naturalness and realism scores for three KC datasets, evaluated by Claude 3.7 Sonnet.}
    \label{tab:natural_realism}
\end{table}

\section{Conclusion}
We propose a KG-based benchmark, MAGIC, for inter-context knowledge conflict detection with greater diversity and complexity.
Experimental results reveal the strengths and limitations of LLMs in handling knowledge conflicts.
Despite recent progress, LLMs continue to struggle with conflict detection in complex cases, e.g., those requiring multi-hop reasoning.
As a future direction, we aim to develop an optimized method to help models overcome these limitations.

\section*{Limitations}
While MAGIC provides a novel benchmark for evaluating knowledge conflict detection, particularly inter-context conflict, it still leaves room for improvement. 
First, MAGIC is built on Wikidata-based knowledge graphs; incorporating additional or domain-specific sources could improve robustness and applicability, and we provide preliminary examples from DBpedia and YAGO in the appendix. 
Future work also includes aligning semantically equivalent relations across graphs to ensure consistency and coverage. 
In addition, since localization evaluation currently relies on human judgment, developing automated approaches could enable a more fine-grained assessment. 

\section*{Ethics Statement}
We release MAGIC, a benchmark dataset for inter-context knowledge conflict detection. 
The dataset is constructed from Wikidata5M, a publicly available resource under permissive licenses, and contains no personally identifiable information. 
While we carefully filtered and validated the generated contexts, we cannot fully guarantee the absence of biased or potentially offensive content. 
MAGIC is released solely for scientific research, with the aim of fostering the development of systems that can reliably handle conflicting information and support safe deployment in real-world applications.

\section*{Acknowledgments}
This work was supported by the National Research Foundation of Korea(NRF) grant funded by the Korea government(MSIT) (RS-2025-00558151). This work was supported by Institute of Information \& communications Technology Planning \& Evaluation (IITP) grant funded by the Korea government(MSIT) (No.RS-2020-II201373, Artificial Intelligence Graduate School Program(Hanyang University)). This work was supported by Institute of Information \& communications Technology Planning \& Evaluation (IITP) under the artificial intelligence semiconductor support program to nurture the best talents (IITP-(2025)-RS-2023-00253914) grant funded by the Korea government(MSIT).

\bibliography{custom}

\clearpage

\appendix


\section{Details of Selected Relation Lists per Domain}
\label{sec:relation_list}
We selected a subset of relations from Wikidata5M based on two criteria: (1) semantic clarity, which enables controlled manipulation, and (2) the ability to form meaningful multi-hop reasoning chains, essential for constructing complex conflict scenarios.

To facilitate more detailed analysis and structured subgraph extraction, we grouped the selected relations into seven semantic domains—\textit{Human, Geography, Organization, Creative Work, Class/Concept, Cause-Effect}, and \textit{General}.
This categorization was manually determined by referring to the official description and subject type constraint of each property on its Wikidata page, based on their inherent meaning and typical usage patterns.

The complete list of selected relations and their domains is shown in Table~\ref{tab:relation_domain}.

\begin{table}[h]
    \centering
    \footnotesize
    \renewcommand{\arraystretch}{1.2}
    \setlength{\tabcolsep}{0.5em}
    \begin{tabular}{|l|p{5.2cm}|}
    \hline
    \textbf{Domain} & \textbf{Relations} \\
    \hline
    Human & P22 (father), P25 (mother), P551 (lived in), P634 (captain), P937 (work location), P1038 (father-in-law), P1066 (student of), P1344 (participant in), P1399 (convicted of), P737 (inflenced by) \\
    \hline
    Geography & P36 (capital), P47 (shares border with), P150 (contains), P189 (find location), P197 (next station), P421 (located in time zone), P1336 (territory claimed by), P3179 (territory overlaps), P1382 (overlaps with), P2789 (connects with) \\
    \hline
    Organization & P127 (owned by), P463 (member of), P807 (separated from), P1001 (belongs to), P2652 (partnership with) \\
    \hline
    Creative Work & P144 (based on), P155 (follows), P178 (developed by), P264 (record label), P287 (designed by) \\
    \hline
    Class/Concept & P279 (subclass of), P460 (equivalent to), P461 (opposite of), P1889 (different from) \\
    \hline
    Cause-Effect & P828 (has cause), P1478 (has immediate cause), P1479 (has contributing factor), P1537 (contributing factor of), P1542 (has result) \\
    \hline
    General & P361 (part of), P527 (consists of), P1011 (excluding), P2283 (uses), P3094 (develops from), P4330 (contains) \\
    \hline
    \end{tabular}
    \caption{Selected relations from Wikidata5M.}
    \label{tab:relation_domain}
\end{table}

\section{Detailed Prompts used in MAGIC}
\paragraph{Prompt for knowledge conflict generation}
Figure~\ref{fig:prompt_kg_generation} shows a prompt used to generate multi-hop conflicts. 
It introduces constraints that guide the LLM to construct indirect contradictions using related subgraphs.
The prompt also defines a clear output format and encourages the use of specific entities and relations in the surrounding subgraphs.
Section~\ref{para:subgraph few shot} mentions how to select few-shot demonstrations used during generation.
For N-conflict cases, the same graph is reused with different perturbations to create multiple conflicts.

\paragraph{Prompt for KG-to-Text conversion}
\label{subsec:kg_to_text_prompt}

We include two prompts used in our KG-to-text pipeline. Figure~\ref{fig:prompt_kg_to_text} prompt guides the generation of natural language contexts from input subgraphs, and Figure~\ref{fig:prompt_kg_to_text_eval} performs automatic verification of triplet coverage using Claude 3.7 Sonnet. Only the outputs that return \textit{No error} in this verification step are retained in our dataset to ensure high-quality generation.

To further guarantee the trustworthiness of the model-based verification, we conduct human inspection on 167 sampled outputs spanning all data types in our dataset. Human annotators evaluate each context using a two-step protocol:
\begin{itemize}[leftmargin=10pt]
    \item \textbf{Conflict Triplet Coverage:} Does the text include the \textit{target} or \textit{perturbed} triplet (i.e., is the intended conflict expressed)?
    \item \textbf{Subgraph Triplet Coverage:} Does the text include \textit{all} subgraph triplets (i.e., is the overall information faithfully conveyed)?
\end{itemize}

As shown in Table~\ref{tab:human_verification}, our method achieves 95.21\% accuracy in the first criterion and 82.04\% in the second.\footnote{With subgraphs often involving more than 10 triplets, achieving over 80\% coverage indicates consistent preservation of essential information.} These results demonstrate that our KG-to-text pipeline is both reliable and automatable, ensuring high-quality generation with minimal manual intervention.

\begin{figure}[h]
    \setlength{\fboxsep}{7pt} 
    \noindent
    \begin{tcolorbox}[
        colback=pink!10,    
        colframe=pink!75,  
        sharp corners=south, 
        rounded corners=northwest,
        boxrule=0.8pt,      
        width=\columnwidth, 
        fonttitle=\bfseries,
        coltitle=black,     
        title=KG-to-Text Conversion Prompt
    ]
        \fontsize{9pt}{11pt}\selectfont 
        \setlength{\parskip}{5pt} 

        \textbf{Instruction} 


        Your task is to convert every provided triplet into a brief, fluent, natural, and coherent single-paragraph in natural language. You MUST include all the facts from the provided triplets. Do NOT omit any triplet or add any information that is not present in the triplets, even if it seems plausible or more natural.
         
    \end{tcolorbox}
    \caption{Prompt for KG-to-Text Conversion.}
    \label{fig:prompt_kg_to_text}
\end{figure}


\begin{figure}[h]
    \setlength{\fboxsep}{7pt} 
    \noindent
    \begin{tcolorbox}[
        colback=pink!10,    
        colframe=pink!75,  
        sharp corners=south, 
        rounded corners=northwest,
        boxrule=0.8pt,      
        width=\columnwidth, 
        fonttitle=\bfseries,
        coltitle=black,     
        title=KG-to-Text Verification Prompt
    ]
        \fontsize{9pt}{11pt}\selectfont 
        \setlength{\parskip}{5pt} 

        \textbf{Instruction} 

    You are an expert KG-to-text error detection system. Your task is to verify whether the provided context faithfully reflects the given set of triplets. You must identify any errors based on the following criteria:


    - \textsc{Incorrect}: The triplet contradicts factual information stated in the context. \\\vspace{-15pt}
    
    - \textsc{Not checkable}: The triplet is not verifiable because the necessary information is missing from the context. \\\vspace{-15pt}
    
    - \textsc{Misleading}: The triplet appears to be present but introduces a misleading or confusing interpretation in the context. \\\vspace{-5pt}

    \textbf{Response Format}
    
    Your response must be one of the following two values only:

    - \textsc{“No error”}: if none of the above errors are present, and the paragraph is concise and fluent. \\\vspace{-15pt}
    
    - \textsc{“Yes error”}: if any of the above errors are present, or if the paragraph is unnaturally verbose or lacks fluency.

    Provide only the answer without any additional explanation.
    
    \end{tcolorbox}
    \caption{Prompt for KG-to-Text verification.}
    \label{fig:prompt_kg_to_text_eval}
\end{figure}

\begin{table}[t]
    \centering
    \small
    \begin{tabular}{l c}
    \toprule
         \textbf{Criteria} & \textbf{Coverage (\%)} \\
    \midrule
        Conflict Triplet & 95.21 \\
        Subgraph Triplet & 82.04 \\
    \bottomrule
    \end{tabular}
    \caption{Human inspection results of KG-to-text outputs based on conflict expression and subgraph coverage.}
    \label{tab:human_verification}
\end{table}

\paragraph{Prompt for knowledge conflict detection}
\label{subsec:binary_prompt}

Figure~\ref{fig:binary_prompt} shows the binary prompt used in \cite{jiayang-etal-2024-econ} for knowledge conflict detection.
In contrast, our stepwise conflict detection prompt is shown in Figure~\ref{fig:multi_step_prompt}, as mentioned in Section~\ref{para:prompting strategy}.

\begin{figure}[h]
    \setlength{\fboxsep}{7pt} 
    \noindent
    \begin{tcolorbox}[
        colback=pink!10,    
        colframe=pink!75,  
        sharp corners=south, 
        rounded corners=northwest,
        boxrule=0.8pt,      
        width=\columnwidth, 
        fonttitle=\bfseries,
        coltitle=black,     
        title=Conflict Detection (Binary Prompt)
    ]
        \fontsize{9pt}{11pt}\selectfont 
        \setlength{\parskip}{5pt} 

        Do the two pieces of context contain conflicting information on answering the question? (Yes/No)

    \end{tcolorbox}
    \caption{Binary prompt for conflict detection.}
    \label{fig:binary_prompt}
\end{figure}

\begin{figure}[t]
    \setlength{\fboxsep}{7pt} 
    \noindent
    \begin{tcolorbox}[
        colback=pink!10,    
        colframe=pink!75,  
        sharp corners=south, 
        rounded corners=northwest,
        boxrule=0.8pt,      
        width=\columnwidth, 
        fonttitle=\bfseries,
        coltitle=black,     
        title=Conflict Detection (Multi-step Prompt)
    ]
        \fontsize{9pt}{11pt}\selectfont 
        \setlength{\parskip}{5pt} 

        \textbf{Instruction} 

        You are given two contexts and your goal is to determine if there are any factual conflicts between them. Ignore what you know and only consider the information within the two contexts.

        \textbf{Response Format} 
        
        If there are no conflicts, output: \texttt{No conflicts} \\\vspace{-15pt}
        
        If there are conflicts, output in this exact format: \\\vspace{-15pt}
        
        Conflicts: \texttt{<number\_of\_conflict>} \\\vspace{-15pt}
        
        Conflict 1: \\\vspace{-15pt}
        
        - Reason: \texttt{<description\_of\_conflict>} \\\vspace{-15pt}
        
        - Sentence A: \texttt{"<sentence\_from\_context A>"} \\\vspace{-15pt}
        
        - Sentence B: \texttt{"<sentence\_from\_context B>"} \\\vspace{-15pt}
        
        .. (Repeat for each conflict)
         
    \end{tcolorbox}
    \caption{Multi-step prompt for conflict detection.}
    \label{fig:multi_step_prompt}
\end{figure}

\paragraph{Prompt for evaluating linguistic quality}

As detailed in Section~\ref{sec:discussion}, we evaluated the linguistic quality of MAGIC contexts along two dimensions: naturalness and realism. 
The detailed evaluation criteria are provided in Figures~\ref{fig:eval_naturalness} and~\ref{fig:eval_realism}.






\begin{figure}[H]
    \setlength{\fboxsep}{7pt} 
    \noindent
    \begin{tcolorbox}[
        colback=pink!10,    
        colframe=pink!75,  
        sharp corners=south, 
        rounded corners=northwest,
        boxrule=0.8pt,      
        width=\columnwidth, 
        fonttitle=\bfseries,
        coltitle=black,     
        title=Context Naturalness Evaluation Prompt
    ]
        \fontsize{9pt}{11pt}\selectfont 
        \setlength{\parskip}{5pt}

    Evaluate the naturalness of the following context on a scale from 0 (very poor) to 5 (excellent).
    
    Focus on: 
    
    - Is the context grammatically correct? \\\vspace{-15pt}
    
    - Does it sound fluent and stylistically natural, as if written by a human? 

    Do not consider factual accuracy or whether the content is realistic. \\\vspace{-15pt}

    Return only a single integer from 0 to 5. Do not provide any explanation or reasoning.
    
    \end{tcolorbox}
    \caption{Prompt for evaluating naturalness of context.}
    \label{fig:eval_naturalness}
\end{figure}

\begin{figure}[H]
    \setlength{\fboxsep}{7pt} 
    \noindent
    \begin{tcolorbox}[
        colback=pink!10,    
        colframe=pink!75,  
        sharp corners=south, 
        rounded corners=northwest,
        boxrule=0.8pt,      
        width=\columnwidth, 
        fonttitle=\bfseries,
        coltitle=black,     
        title=Context Realism Evaluation Prompt
    ]
        \fontsize{9pt}{11pt}\selectfont 
        \setlength{\parskip}{5pt}

    Evaluate the realism of the following context on a scale from 0 (very poor) to 5 (excellent).
    
    Focus on:
    
    - Could this context plausibly occur in a real-world setting? \\\vspace{-15pt}
    
    - Does it resemble something that could realistically appear in natural use cases?

    Do not consider grammar or fluency. Also, do not check whether it is factually accurate. \\\vspace{-15pt}
    
    Return only a single integer from 0 to 5. Do not provide any explanation or reasoning.
    
    \end{tcolorbox}
    \caption{Prompt for evaluating realism of context.}
    \label{fig:eval_realism}
\end{figure}

\section{Human Baseline Performance}
To gain insight into how well humans can perform conflict detection and localization, we conducted a small-scale pilot study.
Specifically, three trained crowd annotators independently reviewed 20 randomly sampled examples from MAGIC each. 

As shown in Table~\ref{tab:human_baseline}, the annotators achieved an average of 92.5\% for conflict identification (ID) and 83.3\% for conflict localization (LOC), generally succeeding in identifying conflicts and their locations with reasonable accuracy.

While these results suggest that humans can reliably detect and localize conflicts--achieving substantially higher accuracy than current models (see Table~\ref{tab:main_table_ID} and Table~\ref{tab:main_table_LOC})--they are based on a small-scale pilot and should not be taken as definitive.
A more rigorous study with a larger and more diverse annotator pool will be conducted in future work.

\begin{table}[h]
    \centering
    \small
    \begin{tabular}{l c c}
    \toprule
         \textbf{Method} & \textbf{ID (\%)} & \textbf{LOC (\%)} \\
    \midrule
        Human Performance & 92.5 & 83.3 \\
    \bottomrule
    \end{tabular}
    \caption{Human performance on the identification and localization evaluated on a small subset of MAGIC.}
    \label{tab:human_baseline}
\end{table}

\section{Detailed Model Evaluation on MAGIC}
\subsection{Performance by each Conflict type}
\label{subsec:detailed_results}
Table~\ref{tab:detailed_ID} and Table~\ref{tab:detailed_LOC} provide detailed ID and LOC scores for five models across different conflict types in MAGIC dataset.
Table~\ref{tab:detailed_ID} includes smaller model performance, such as Mistral-7B-Instruct-v0.1~\cite{mistral2023instruct} and Llama 3.1 8B Instruct~\cite{dubey2024llama}.
As discussed in Section~\ref{sec:main results}, when the number of conflicts (N) grows, ID scores tend to increase while LOC scores decrease.
Multi-Hop conflicts generally yield lower performance in both ID and LOC compared to Single-Hop cases.


\begin{table*}
    \centering
    \footnotesize
    \setlength{\tabcolsep}{0.6em}
    \begin{tabular}{l >{\columncolor{gray!10}}c >{\columncolor{gray!10}}c >{\columncolor{gray!10}}c >{\columncolor{gray!10}}c c c c c c}
    \toprule
        \multirow{2}{*}{\shortstack[c]{\textbf{}}} 
        & \multicolumn{4}{>{\columncolor{gray!10}}c}{\textbf{Single-Hop}} & \multicolumn{4}{c}{\textbf{Multi-Hop}} \\
        & N=1 & N=2 & N=3 & N=4 & N=1 & N=2 & N=3 & N=4 \\
    \midrule
        Mixtral 8x7B & 42.72 & 51.66 & 51.90 & 67.35 & 23.47 & 31.61 & 38.16 & 30.00 \\
        Llama 3.1 70B & 71.43 & 79.31 & 93.88 & 90.48 & 59.52 & 78.67 & 70.00 & 73.91 \\
        Claude 3.5 Haiku & 67.79 & 72.73 & 86.25 & 84.00 & 41.00 & 44.30 & 63.75 & 86.00 \\
        GPT-4o-mini & 85.58 & 87.66 & 100.00 & 98.00 & 70.67 & 84.18 & 86.25 & 94.00 \\
        o1 & 87.02 & 90.26 & 97.50 & 98.00 & 36.00 & 58.23 & 71.25 & 62.00 \\
        \midrule
        Average & 70.91 & 76.32 & 85.91 & 87.57 & 46.13 & 59.40 & 65.88 & 69.18 \\
        \midrule
        Mistral 7B & 7.43 & 11.82 & 7.41 & 10.81 & 11.06 & 6.93 & 4.00 & 12.90 \\
        Llama 3.1 8B & 7.21 & 15.59 & 23.00 & 24.00 & 6.33 & 13.38 & 10.00 & 12.00 \\
    \bottomrule
    \end{tabular}
    \caption{ID score by model on MAGIC.}
    \label{tab:detailed_ID}
\end{table*}

\begin{table*}
    \centering
    \footnotesize
    \setlength{\tabcolsep}{0.6em}
    \begin{tabular}{l >{\columncolor{gray!10}}c >{\columncolor{gray!10}}c >{\columncolor{gray!10}}c >{\columncolor{gray!10}}c c c c c c}
    \toprule
        \multirow{2}{*}{\shortstack[c]{\textbf{}}} 
        & \multicolumn{4}{>{\columncolor{gray!10}}c}{\textbf{Single-Hop}} & \multicolumn{4}{c}{\textbf{Multi-Hop}} \\
        & N=1 & N=2 & N=3 & N=4 & N=1 & N=2 & N=3 & N=4 \\
    \midrule
        Mixtral 8x7B & 38.83 & 21.85 & 15.19 & 14.29 & 12.59 & 7.10 & 6.58 & 0.00 \\
        Llama 3.1 70B & 62.64 & 42.53 & 40.82 & 42.86 & 31.75 & 25.33 & 25.00 & 8.70 \\
        Claude 3.5 Haiku & 48.08 & 57.79 & 57.50 & 48.00 & 33.67 & 35.44 & 33.75 & 32.00 \\
        GPT-4o-mini & 78.85 & 53.90 & 58.75 & 44.00 & 54.67 & 47.47 & 33.75 & 24.00 \\
        o1 & 86.54 & 66.88 & 68.75 & 62.00 & 30.67 & 30.38 & 27.50 & 12.00 \\
        \midrule
        Average & 62.99 & 48.59 & 48.20 & 42.23 & 32.67 & 29.14 & 25.32 & 15.34 \\
    \bottomrule
    \end{tabular}
    \caption{LOC score by model on MAGIC.}
    \label{tab:detailed_LOC}
\end{table*}

\subsection{Domain-Level Analysis}
\label{subsec:perform_domain_rel}
Table~\ref{tab:perform_by_domain} presents the average ID scores of models across seven domains, and Figure~\ref{fig:perform_by_relation} shows the averaged ID scores of models across relation types.
These results are under the 1-conflict setting in MAGIC.
For each model, the highest score is highlighted in red, while the lowest is in green.

\begin{figure*}[h]
\centering
\includegraphics[width=0.9\textwidth]{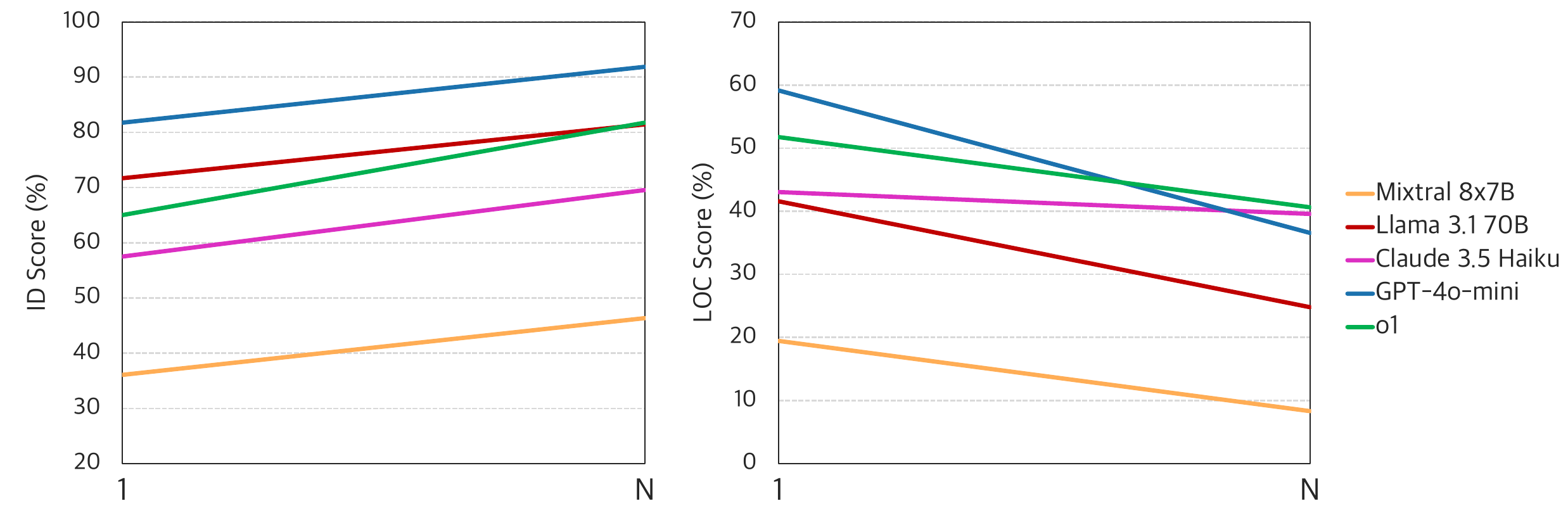}
  \caption{Performance by number of domains per sample in MAGIC.}
\label{fig:perform_by_number_of_domain}
\end{figure*}

Figure~\ref{fig:perform_by_number_of_domain} shows the performance based on the number of domains included in each data instance.
The results indicate that as the number of domains increases, ID scores tend to improve, while LOC scores decrease, showing that models more easily distinguish conflicts in multi-domain settings but struggle with precise localization.


\begin{table}
    \centering
    \footnotesize
    \setlength{\tabcolsep}{0.2em}
    \begin{tabular}{l c c c c c c c c c c}
    \toprule
        \textbf{Domain} & \textbf{Mixtral} & \textbf{Llama} & \textbf{Haiku} & \textbf{GPT-4o-m} & \textbf{o1} \\
    \midrule
        Human & 22.64 & 56.52 & 54.55 & {\cellcolor{green!30}}69.09 & 63.64 \\
        Geography & 25.95 & 72.73 & 52.20 & 79.87 & 56.60 \\
        Organization & {\cellcolor{green!30}}14.52 & {\cellcolor{red!30}}75.86 & {\cellcolor{green!30}}42.86 & 74.60 & 42.86 \\
        Creative Work & 37.93 & 66.67 & 55.17 & {\cellcolor{red!30}}82.76 & {\cellcolor{green!30}}41.38 \\
        Class/Concept & {\cellcolor{red!30}}47.89 & 68.97 & {\cellcolor{red!30}}55.41 & 75.68 & {\cellcolor{red!30}}72.97 \\
        Cause-Effect & 32.56 & 54.55 & 46.51 & 72.09 & 53.49 \\
        General & 42.86 & {\cellcolor{green!30}}47.22 & 55.29 & 78.82 & 56.47 \\
    \bottomrule
    \end{tabular}
    \caption{Domain-level analysis with 1-conflict problems in MAGIC.}
    \label{tab:perform_by_domain}
\end{table}

\section{Performance by Number of Conflicts}
\label{sec:econ_multiple}


\begin{figure}[h]
    \includegraphics[width=0.9\linewidth]{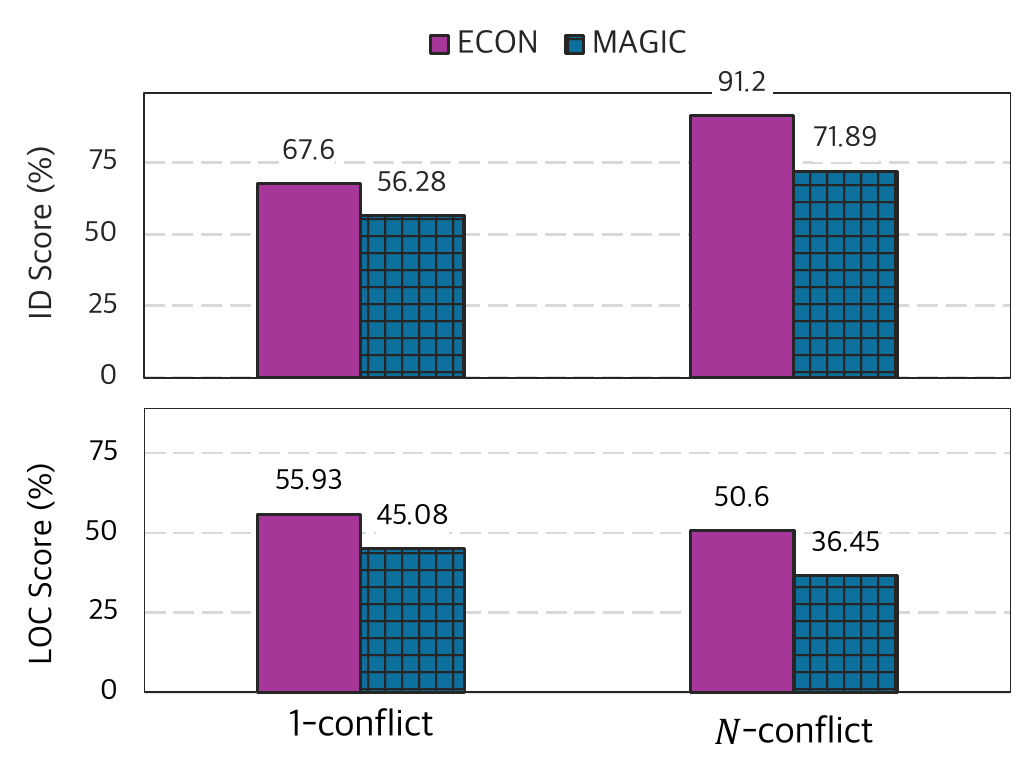}
  \caption{Comparison of detection performance by number of conflicts. ECON’s factoid conflicts contain multiple (N) conflicts that span across sentences.}
  \label{fig:compare_econ_multiple}
\end{figure}

Figure \ref{fig:compare_econ_multiple} compares performance across ECON and MAGIC based on the number of conflicts (1-conflict vs. N-conflict).
Note that ECON’s factoid conflicts involve multiple conflicts introduced across several sentences.
This aligns with our findings, suggesting that while a higher number of conflicts facilitates conflict identification, it also makes precise localization more challenging.
Conversely, when multiple conflicts occur, identifying all specific conflicting sentences becomes more difficult, leading to a decrease in the LOC score.

\section{Impact of Subgraph Scope on Conflict Generation}
\label{sec:triplet_subgraph}
\label{para:preliminary}
In a preliminary attempt, we prompted the model to generate conflicts using only a selected seed triplet, without incorporating the surrounding subgraph.
Table~\ref{tab:triplet_example} shows that this often resulted in trivial patterns—such as simply negating the original relation or replacing with incoherent, off-topic facts. 
These observations underscore the importance of incorporating broader context, as our subgraph-level prompting enables the generation of more realistic and semantically grounded conflicts.

\begin{table*}[t]
    \centering
    \small
    \begin{tabular}{l p{0.8\textwidth}}
    \toprule
        \multicolumn{2}{>{\columncolor{gray!10}}c}{\textbf{1-Multi-Hop}} \\
    \midrule
        Original Triplet & (Moskva | contains | staroye kryukovo district) \\
    \midrule
        Perturbed Triplet & (Moskva | borders | Odintsovo), (Odintsovo | contains | staroye kryukovo district) \\
    \midrule
        Subgraph & (Moskva | contains | kosino-ukhtomsky district), (Moskva | contains | Prospekt Vernadskogo District), (Moskva | divides into | Chertanovo Tsentralnoye District), (Moskva | twinned administrative body | tunis)\\
    \midrule
        Context1 & Moskva is a city that contains several districts, including the staroye kryukovo district, kosino-ukhtomsky district, and Prospekt Vernadskogo District. Additionally, it is divided into Chertanovo Tsentralnoye District. Moskva also has a twinned administrative body relationship with Tunis. \\
    \midrule
        Context2 & Moskva borders Odintsovo and contains several districts, including the Kosino-Ukhtomsky District, the Prospekt Vernadskogo District, and it also divides into Chertanovo Tsentralnoye District. Additionally, Odintsovo contains the Staroye Kryukovo District, and Moskva is twinned with the administrative body of Tunis.\\
    \midrule
        \multicolumn{2}{>{\columncolor{gray!10}}c}{\textbf{2-Single-Hop}} \\
    \midrule
        Original Triplet \#1 & (Hastings, New Brunswick | territory overlaps | Kings County, New Brunswick) \\
    \midrule
        Perturbed Triplet \#1 & (Hastings, New Brunswick | territory does not overlap | Kings County, New Brunswick) \\
    \midrule
        Original Triplet \#2 & (Hastings, New Brunswick | territory overlaps | albert county) \\
    \midrule
        Perturbed Triplet \#2 & (Hastings, New Brunswick | territory does not overlap | albert county) \\
    \midrule
        Subgraph & (Hastings, New Brunswick | instance of | a dark-sky preserve), (Hastings, New Brunswick | operator | canadian parks service), (Hastings, New Brunswick | member of | Canadian National Parks), (Canadian National Parks | subclass of | national park), (Canadian National Parks | has list | List of national parks of Canada), (Canadian National Parks | subclass of | Protected areas of Canada), (albert county | located in the administrative territorial entity | Culture of New Brunswick), (albert county | shares border with | saint john county) \\
    \midrule
        Context1 & Hastings, New Brunswick, is a dark-sky preserve operated by the Canadian Parks Service and is a member of Canadian National Parks, which is a subclass of national parks and protected areas in Canada. The territory of Hastings overlaps with Kings County and Albert County, the latter of which is located within the administrative territorial entity of the Culture of New Brunswick and shares a border with Saint John County. \\
    \midrule
        Context2 & Hastings, New Brunswick, is recognized as a dark-sky preserve and is operated by the Canadian Parks Service, making it a member of the Canadian National Parks, which is a subclass of national parks and protected areas in Canada. The territory of Hastings does not overlap with Kings County or Albert County, the latter of which is situated within the administrative territorial entity of the Culture of New Brunswick and shares a border with Saint John County. The Canadian National Parks also maintains a list known as the List of national parks of Canada. \\
    \bottomrule
    \end{tabular}
    \caption{Examples of triplet-level dataset generation.}
    \label{tab:triplet_example}
\end{table*}

\section{Annotation Guideline}
\label{sec:annotation_guideline}

To ensure the quality of our MAGIC dataset, human intervention was applied at two stages of data construction pipeline: (1) manually selecting of few-shot demonstrations used for prompting, and (2) filtering out trivial or inherent model-generated conflicts after generation.

All annotations were performed independently by two researchers following a shared guideline to ensure consistency across relation types and conflict settings.
The detailed annotation guidelines for each stage are provided below.

\setlength{\fboxsep}{7pt}
\noindent
\begin{tcolorbox}[
  colback=gray!3,           
  colframe=black!80,        
  boxrule=0.8pt,            
  arc=2pt,                  
  width=\columnwidth,       
  fonttitle=\bfseries,      
  coltitle=black,           
  left=5pt,                 
  right=5pt,                
  top=5pt,                  
  bottom=5pt                
]

\fontsize{9pt}{11pt}\selectfont
\setlength{\parskip}{5pt}

\textbf{Guideline for Few-shot Demonstration Selection}

The goal of this task is to select three three representative examples per relation type to be included in the few-shot prompt.
These examples should be chosen from zero-shot model generations and must serve as effective demonstrations of plausible and challenging knowledge conflicts.

Selected examples should follow the criteria below:

\begin{itemize}[leftmargin=10pt]
    \item Each example must express a plausible and semantically coherent knowledge conflict.
    \item The conflict must be appropriate for the given relation, reflecting its intended usage in Wikidata.
    \item Examples involving multi-hop reasoning or indirect contradictions were preferred over surface-level entity substitutions.
    \item Redundant or structurally repetitive patterns across examples were avoided to ensure diversity.
\end{itemize}
\end{tcolorbox}

\setlength{\fboxsep}{7pt}
\noindent
\begin{tcolorbox}[
  colback=gray!3,           
  colframe=black!80,         
  boxrule=0.8pt,            
  arc=2pt,                  
  width=\columnwidth,       
  fonttitle=\bfseries,      
  coltitle=black,           
  left=5pt,                 
  right=5pt,                
  top=5pt,                  
  bottom=5pt                
]

\fontsize{9pt}{11pt}\selectfont
\setlength{\parskip}{5pt}

\textbf{Guideline for Post-generation Conflict Filtering}

The goal of this task is to identify and remove low-quality outputs from model-generated conflict instances. 
Annotators should review each instance generated via prompting and apply the following criteria to filter out unsuitable samples:

\begin{itemize}[leftmargin=10pt]
    \item For single-hop conflicts, the perturbed triplet must contradict the original triplet in a direct and unambiguous manner.
    \item For multi-hop cases, the contradiction must emerge through a reasoning chain spanning multiple triplets.
    \item In N-conflict instances, each conflict must be logically independent and non-overlapping with others in the same context pair.
    \item Outputs with unnatural phrasing, semantic incoherence, or implausible context were discarded.

\end{itemize}
\end{tcolorbox}

\section{The Scalability and Consistency of Human Evaluation}
\label{sec:human eval}
\paragraph{Necessity of human evaluation}
Current LLMs—including those we benchmarked—are not yet reliable for conflict detection and localization (see Tables~\ref{tab:main_table_ID} and \ref{tab:main_table_LOC}). 
Therefore, human evaluation was necessary and inevitable.

\paragraph{Efficient human evaluation protocol}
To reduce the annotation cost, we adopted the following efficient protocol.  
First, since every data instance in MAGIC is designed to contain at least one conflict, predictions of "No conflict" can be safely ignored as they are necessarily incorrect.  
Second, localization (identifying conflicting spans) was evaluated only when identification (existence of conflict) was correct.  
Finally, to check the correctness of localization, we reviewed results in bulk by comparing outputs from multiple models and gold-standard labels in parallel.
Applying these heuristics significantly reduced the annotation workload, eliminating the need to evaluate each case individually.

\paragraph{Attempts at automatic evaluation}
We also explored the feasibility of automating evaluation. 
Specifically, we conducted an experiment on 150 randomly sampled examples, selecting GPT-4o-mini, which demonstrated the best localization performance, as the evaluation model to assess whether conflict localization predictions from other models are correct.  
In other words, GPT-4o-mini served as an automatic judge to evaluate the localization accuracy of o1 and Mixtral-8x7B.

We used the following prompt in Figure~\ref{fig:automatic_prompt} with GPT-4o-mini, which clearly specifies the steps for conflict localization and verification:

\begin{figure}[h]
    \setlength{\fboxsep}{7pt} 
    \noindent
    \begin{tcolorbox}[
        colback=pink!10,    
        colframe=pink!75,  
        sharp corners=south, 
        rounded corners=northwest,
        boxrule=0.8pt,      
        width=\columnwidth, 
        fonttitle=\bfseries,
        coltitle=black,     
        title=Automatic LOC Evaluation Prompt
    ]
        \fontsize{9pt}{11pt}\selectfont 
        \setlength{\parskip}{5pt}

    You are an expert evaluator of knowledge conflict analysis. Below are two documents: 

    - Document A: \texttt{\{docA\}}
    \\\vspace{-15pt}
    
    - Document B: \texttt{\{docB\}}
    
    The model predicted the following conflicting spans: 
    
    - Span A: \texttt{\{sentences[0]\}}
    \\\vspace{-15pt}
    
    - Span B: \texttt{\{sentences[1]\}}

    Your tasks:
    
    1) For each span, identify the exact sentence from its document that contains the span.
    \\\vspace{-15pt}
    
    2) Determine whether the two sentences identified in step 1 truly contradict each other in the context of both documents.
    \\\vspace{-15pt}
    
    3) Answer "Yes" if they exactly match sentences in their respective documents and the sentences contradict each other, or "No" if they do not.
    
    Provide only "Yes" or "No" without any explanation.
    
    \end{tcolorbox}
    \caption{Prompt used for automatic conflict localization evaluation.}
    \label{fig:automatic_prompt}
\end{figure}

\paragraph{Results}
Table~\ref{tab:automatic_LOC} reports the accuracy of GPT-4o-mini’s evaluations compared to human annotations.

As shown, GPT-4o-mini achieves only 74.67\% agreement with human annotations for o1 and 48.72\% for Mixtral-8x7B. 
These results highlight that, without human effort, current LLMs still fall short of achieving performance sufficient to reliably replace human judgement in conflict localization.


\begin{table}[t]
    \centering
    \small
    \begin{tabular}{l c c}
    \toprule
         \textbf{Model being tested} & \textbf{o1} & \textbf{Mixtral 8x7B} \\
    \midrule
        \makecell{LOC score \\ \scriptsize (by GPT-4o-mini)} 
            & 74.67 & 48.72 \\
    \bottomrule
    \end{tabular}
    \caption{LOC scores of model outputs evaluated by GPT-4o-mini (relative to human annotations set to 100\%).}
    \label{tab:automatic_LOC}
\end{table}


\section{The Generalizability of the Benchmark Framework}
Our framework is not limited to Wikidata5M but can be directly applied to other knowledge graphs. 
To verify this, we experimented with DBpedia \cite{doi:10.3233/SW-140134} and YAGO \cite{suchanek:hal-01472497} and observed substantial overlap between their relation sets and the relation pool defined in our framework (e.g., \textit{child}, \textit{capital}, \textit{participatedIn}). 

Table~\ref{tab:other_kgs} presents representative examples generated from DBpedia and YAGO using the same pipeline, confirming the compatibility and extensibility of our approach. 
These results highlight that our benchmark construction method is robust and readily generalizable to diverse knowledge graphs beyond Wikidata5M.

\begin{table*}[t]
    \centering
    \small
    \begin{tabular}{l p{0.8\textwidth}}
    \toprule
         \multicolumn{2}{>{\columncolor{gray!10}}c}{\textbf{DBpedia}} \\
    \midrule
        Original Triplet & (Albennie\_Jones | genre | Jazz), (Jazz | derivative | Funk) \\
    \midrule
        Perturbed Triiplet & (Albennie\_Jones | opposite | Funk) \\
    \midrule
        Subgraph & (Albennie\_Jones | activeYearsStartYear | 1930), (Albennie\_Jones | background | "solo\_singer") (Ace\_Wilder | background | "solo\_singer") \\
    \midrule
        Context1 &	lbennie Jones was a solo singer who began performing in 1930. His main genre was Jazz, which later evolved into Funk. Ace Wilder is also noted as a solo singer. \\
    \midrule
        Context2 & Albennie Jones was a solo singer who started his career in 1930. However, his musical style was completely opposite to Funk. Another solo singer mentioned here is Ace Wilder. \\
    \midrule
         \multicolumn{2}{>{\columncolor{gray!10}}c}{\textbf{YAGO}} \\
    \midrule
        Original Triplet & (Monrovia | isLocatedIn | Africa) \\
    \midrule
        Perturbed Triplet & (Monrovia | isLocatedIn | China)\\
    \midrule
        Subgraph & (Liberia | dealsWith | France), (Liberia | hasCapital | Monrovia), (Tang\_dynasty | hasCapital | Luoyang), (Luoyang | isLocatedIn | China) \\
    \midrule
        Context1 & Liberia, an African country, has its capital in Monrovia, which is located in Africa. Liberia maintains diplomatic relations with France. In a different historical context, the Tang dynasty had its capital in Luoyang, a city situated in China. \\
    \midrule
        Context2 & Liberia, a country whose capital is Monrovia, maintains diplomatic relations with France. Interestingly, Monrovia is described as being located in China. During the Tang dynasty, the capital city was Luoyang, which, like Monrovia, is located in China. \\
    \bottomrule
    \end{tabular}
    \caption{Conflict examples constructed from different KG datasets, DBpedia and YAGO.}
    \label{tab:other_kgs}
\end{table*}

\section{Examples from MAGIC Dataset}
\label{sec:example_magic}

Table~\ref{tab:ex_single_hop} (Single-Hop) and Table~\ref{tab:ex_multi_hop} (Multi-Hop) show example contexts from MAGIC dataset.

\begin{table*}[t]
    \centering
    \small
    \begin{tabular}{l p{0.8\textwidth}}
    \toprule
         \multicolumn{2}{>{\columncolor{gray!10}}c}{\textbf{1-Single-Hop}} \\
    \midrule
        Context1 & Guy Williams, a basketball player, is distinct from Gus Williams, another basketball player. He has been a member of the Baltimore Bullets and the Oakland Warriors during his career. It is important to note that he is different from James “Fly” Williams. The name Guy, which is his given name, is equivalent to the name Guido and belongs to the French vocabulary. Additionally, the name Guy is also a surname that is identical to the given name. The writing system used for the name Guy is Latin alphabet letters, which are based on the roman-alphabet and are an instance of an alphabetic writing system. The history of the Latin alphabet is the historical context surrounding the use of these letters. \\
        Context2 & Guy Williams, a basketball player, is the same person as Gus Williams, who is also known as a basketball player. The name Guy is equivalent to the given name Guido, is of French vocabulary origin, and shares a family name identical to Guy (surname). The writing system for the name Guy is Latin alphabet letters, which are instances of an alphabetic writing system based on the Roman alphabet. Latin alphabet letters have a historical context in the history of the Latin alphabet. In his basketball career, Guy Williams was a member of the Baltimore Bullets and the Oakland Warriors, and he is different from another player named James “Fly” Williams. \\
    \midrule
        \multicolumn{2}{>{\columncolor{gray!10}}c}{\textbf{2-Single-Hop}} \\
    \midrule
        Context1 & The concept of the "Margin of opportunity" overlaps with both the "Sensitive period" and the "Time limit," and is classified as a subclass of the broader category of "event." This margin is also a facet of both "WikiProject Urban studies" and "Orbital maneuver." Within the realm of knowledge, "mastery" is seen as a subclass of "Knowledgeableness" and "aptitude," and is described by the source known as "el panson." Additionally, the term "event" is used by a "Relativistic observer" and is equivalent to "Event (statistics)." Notably, "WikiProject Urban studies" itself falls under the subclass of "mastery." \\
        Context2 & The Margin of Opportunity is disjoint from the Sensitive Period and does not overlap with the Time Limit. It is considered a subclass of events and a facet of both WikiProject Urban Studies and Orbital Maneuver. Mastery is a subclass of both Knowledgeableness and Aptitude, and is described by the source El Panson. The concept of an event is used by a relativistic observer and is equivalent to an event in statistics. Lastly, WikiProject Urban Studies is a subclass of mastery. \\
    \midrule
        \multicolumn{2}{>{\columncolor{gray!10}}c}{\textbf{3-Single-Hop}} \\
    \midrule
        Context1 & Bilecik University is a public college located in Bilejik and has separated from several institutions, including Kütahya Dumlupınar University, Eskişehir Osmangazi University, and Anadolu University. \\
        Context2 & Bilecik University is a public college located in Bilejik and has merged with several institutions, including Kütahya Dumlupınar University, Eskişehir Osmangazi University, and Anadolu University. \\
    \midrule
        \multicolumn{2}{>{\columncolor{gray!10}}c}{\textbf{4-Single-Hop}} \\
    \midrule
        Context1 & John, a personal name, is equivalent to Ifan in some contexts and can also be represented as Jean and Ioannis in different languages. In addition, the name John has a specific connection to John Hervey, who lived from 1616 to 1680 and was the father-in-law of Robert Jermyn. John Hervey was involved in significant historical events, as he was a member of the Royal Society of Great Britain and the Restoration Parliament, and he lived during the Civil War in England. The name John is primarily associated with the German language (iso 639:deu), which is classified as a High German dialect, has a V2 word order, possesses a simple will future tense, and includes grammatical cases such as the genitive case. \\
        Context2 & John is a personal name that is not equivalent to the given name Ifan, but it is equivalent to the names Hans and Johannes. The German language, denoted as iso 639:deu, is a High German language characterized by v2 word order, the simple will future ii tense, and the genitive case. John Hervey, who lived from 1616 to 1680, was the son-in-law of Robert Jermyn and was involved in the Civil War in England. He was also a member of the Royal Society of Great Britain and the Restoration Parliament. \\
    \bottomrule
    \end{tabular}
    \caption{Examples for Single-Hop conflict in MAGIC.}
    \label{tab:ex_single_hop}
\end{table*}

\begin{table*}[t]
    \centering
    \small
    \begin{tabular}{l p{0.8\textwidth}}
    \toprule
         \multicolumn{2}{>{\columncolor{gray!10}}c}{\textbf{1-Multi-Hop}} \\
    \midrule
        Context1 & Hastings, New Brunswick, is an area that overlaps with Saint John County and Albert County, and it is recognized as a dark-sky preserve, an instance of a terrestrial protected area. This preserve, which is part of the Fundy Biosphere Reserve, is under the operation of the Canadian Parks Service and has been conferred the designation of a dark-sky preserve by both the Dark Sky Association and the Bulletin of the Royal Astronomical Society of Canada. The Dark Sky Association, an environmental organization focused on creating darker skies through initiatives like "lights out for darker skies," has its field of work centered on this ecological effort and is also located within the administrative territorial entity of Satori Charter School. \\
        Context2 & Hastings, located in New Brunswick, overlaps with Albert County and is part of the Fundy Biosphere Reserve, operated by the Canadian Parks Service. This area is recognized as a dark-sky preserve, a designation conferred by both the Bulletin of the Royal Astronomical Society of Canada and the Dark Sky Association, which is an environmental organization focused on promoting darker skies through initiatives like "Lights Out for Darker Skies." It is important to note that Albert County is completely disjoint from Saint John County. \\
    \midrule
        \multicolumn{2}{>{\columncolor{gray!10}}c}{\textbf{2-Multi-Hop}} \\
    \midrule
        Context1 & Auitzotl was the son of Atotoztli II and Epcoatl. He had a sibling named Acolnahuacatl Tzacualcoatl and was the father of two children, Chimalpilli II and Cuahatemoc. \\
        Context2 & Auitzotl is the parent of Cuahatemoc and Chimalpilli II, and has a sibling named Acolnahuacatl Tzacualcoatl, who is the child of Epcoatl. Cuahatemoc's mother is Atotoztli II. \\
    \midrule
        \multicolumn{2}{>{\columncolor{gray!10}}c}{\textbf{3-Multi-Hop}} \\
    \midrule
        Context1 & The State Penn is a member of several organizations, including the Digital Library Federation, SPARC Europe, and the Center for Research Libraries (CRL). Additionally, it is affiliated with the Oak Ridge Associated Universities, which is located in Oak Ridge, Tennessee, and operates as a matrix organization. Otto Poggeler, who serves as an employee at State Penn, was born in Attendorn, Germany, and is a member of the North Rhine-Westphalia Academy for Sciences and Arts. He speaks and writes in German, known by the ISO 639 code as deu. \\
        Context2 & State Penn is a member of Oak Ridge Associated Universities, which is a matrix organization located in Oak Ridge, Tennessee. The Digital Library Federation excludes membership for matrix organizations, and similarly, SPARC Europe restricts membership to institutions located in Europe. Oak Ridge Associated Universities is also considered a matrix organization, which is incongruent with the Center for Research Libraries (CRL). In addition, Otto Poggeler, who was born in Attendorn, Germany, speaks German and is a member of the North Rhine-Westphalia Academy for Sciences and Arts. He is employed by State Penn. \\
    \midrule
        \multicolumn{2}{>{\columncolor{gray!10}}c}{\textbf{4-Multi-Hop}} \\
    \midrule
        Context1 & The name "Iulian" is equivalent to several other names including "Julian," "Julio," "Juliusz," and "Julien." In Modern Spanish, the name "Julián" serves as its counterpart and is also equivalent to "Jules," "Julien," and "Juliusz," while "Iulian" further connects to "Julián" as a first name. Additionally, Modern Spanish is classified under the Castilian languages and features various grammatical moods and tenses, including the conditional tense, present indefinite tense, and past perfect simple. \\
        Context2 & The name Julián is a given name in Modern Spanish, equivalent to several other names including iulian, Juliusz, jules, and Julien, though it is distinct from the name julian. The origins of the name Julio can be traced back to Latin, while Julián's various equivalents reflect its connections across different languages and cultures. All references to Julián confirm its continuous usage in Modern Spanish, which is a subclass of Castilian languages characterized by grammatical moods such as the conditional tense and various tenses including the present indefinite and past perfect simple. \\
    \bottomrule
    \end{tabular}
    \caption{Examples for Multi-Hop conflict in MAGIC.}
    \label{tab:ex_multi_hop}
\end{table*}

\subsection{Qualitative Analysis of Model Failures}

Individual failures reveal deeper challenges. 
GPT-4o-mini, despite its overall strength, struggles with multi-hop reasoning over densely connected subgraphs. 
As shown in the below of Table~\ref{tab:easy_hard_example}, one MAGIC includes a case where \textit{"John is equivalent to Hans"}, while the perturbed ones ultimately imply that \textit{John is not equivalent to Hans"}.
Detecting this contradiction requires chaining equivalence and distinction relations across multiple entities.

\begin{table*}[t]
    \centering
    \small
    \begin{tabular}{l p{0.8\textwidth}}
    \toprule
    \multicolumn{2}{>{\columncolor{gray!10}}c}{\textbf{Easy Example}} \\
    \midrule
        Original Triplet & (1891 British Lions tour to South Africa | captain | bill maclagen) \\
    \midrule
        Perturbed Triplet & (1891 British Lions tour to South Africa | captain | william burrows) \\
    \midrule
        Subgraph & (1891 British Lions tour to South Africa | destination point | Suid Africa), (1891 British Lions tour to South Africa | follows | 1888 british lions tour to new zealand and australia), (1891 British Lions tour to South Africa | sport | Fifteen-a-side), (1891 British Lions tour to South Africa | followed by | 1896 British Lions tour to South Africa), (Suid Africa | diplomatic relation | hellada), (Suid Africa | part of | Continent of Africa), (Suid Africa | diplomatic relation | argentina), (1896 British Lions tour to South Africa | followed by | 1899 british lions tour to australia), (1896 British Lions tour to South Africa | destination point | Suid Africa) \\
    \midrule
        Context1 & The 1891 British Lions tour to South Africa, captained by Bill Maclagen, was a fifteen-a-side rugby tour that took place in Suid Africa, following the 1888 British Lions tour to New Zealand and Australia. This tour was later followed by the 1896 British Lions tour to South Africa, which also had Suid Africa as its destination point. Suid Africa is located on the Continent of Africa and maintains diplomatic relations with Hellada and Argentina. The 1896 tour would subsequently be followed by the 1899 British Lions tour to Australia. \\
    \midrule
        Context2 & The 1891 British Lions tour to South Africa, captained by William Burrows, was a Fifteen-a-side rugby tour that took place in Suid Africa, following the earlier 1888 British Lions tour to New Zealand and Australia. This tour was succeeded by the 1896 British Lions tour to South Africa, which also had Suid Africa as its destination point. Suid Africa, part of the Continent of Africa, has diplomatic relations with Hellada and Argentina. The 1896 British Lions tour to South Africa was subsequently followed by the 1899 British Lions tour to Australia. \\
    \midrule
        \multicolumn{2}{>{\columncolor{gray!10}}c}{\textbf{Difficult Example 1}} \\
    \midrule
        Original Triplet & (Henry Charles Fitzroy Somerset, 8th Duke of Beaufort | work location | united kingdom/london) \\
    \midrule
        Perturbed Triplet & (Henry Charles Fitzroy Somerset, 8th Duke of Beaufort | place of death | Stoke Gifford), (Stoke Gifford | is part of | South Gloucestershire), (South Gloucestershire | is geographically distinct from | united kingdom/london) \\
    \midrule
        Subgraph & (Henry Charles Fitzroy Somerset, 8th Duke of Beaufort | position held | uk mp), (Henry Charles Fitzroy Somerset, 8th Duke of Beaufort | place of death | Stoke Gifford), (Henry Charles Fitzroy Somerset, 8th Duke of Beaufort | member of political party | new tories), (Henry Charles Fitzroy Somerset, 8th Duke of Beaufort | child | Blanche Scott Douglas) \\
    \midrule
        Context1 & Henry Charles Fitzroy Somerset, the 8th Duke of Beaufort, worked in London, United Kingdom, and held the position of MP for the UK as part of the New Tories political party. He passed away in Stoke Gifford, and is also known to have had a child named Blanche Scott Douglas. \\
    \midrule
        Context2 & Henry Charles Fitzroy Somerset, the 8th Duke of Beaufort, passed away in Stoke Gifford, a locality that is part of South Gloucestershire, which is geographically distinct from London in the United Kingdom. He held the position of a UK Member of Parliament and was a member of the New Tories political party. He is also the father of Blanche Scott Douglas. \\
    \midrule
        \multicolumn{2}{>{\columncolor{gray!10}}c}{\textbf{Difficult Example 2}} \\
    \midrule
        Original Triplet & (Jon (first name) | equivalent to | Hans (first name)) \\
    \midrule
        Perturbed Triplet & (Jon (first name) | equivalent to | gianni (first name)), (gianni (first name) | equivalent to | Ivan), (Ivan | different from | Hans (first name)) \\
    \midrule
        Subgraph & (gianni (first name) | equivalent to | jaan (first name)), (gianni (first name) | equivalent to | Ivan), (gianni (first name) | equivalent to | juan (first name)), (gianni (first name) | equivalent to | Ioannis (name)), (Jon (first name) | equivalent to | Evan), (Jon (first name) | equivalent to | jaan (first name)), (Jon (first name) | equivalent to | seán), (Jon (first name) | equivalent to | gianni (first name)), (Ioannis (name) | equivalent to | ifan (given name)), (Ioannis (name) | equivalent to | ion (given name)), (Ioannis (name) | equivalent to | Jean (first name)), (Ioannis (name) | equivalent to | johnny (first name)) \\
    \midrule
        Context1 & The name Jon is equivalent to several other names, including Hans, Evan, jaan, and seán. Additionally, Jon is also synonymous with Gianni, which itself is equivalent to Jaan, Ivan, Juan, and Ioannis. Ioannis can further be linked to Ifan, Ion, Jean, and Johnny, showcasing a web of connections among these various names. \\
    \midrule
        Context2 & The name Gianni is equivalent to several other names, including Ivan, Jaan, Juan, and Ioannis. Additionally, it is noted that Ivan is different from Hans. Jon is another name that shares equivalencies, as it is equivalent to Evan, Jaan, Seán, and Gianni. Furthermore, the name Ioannis is equivalent to Ifan, Ion, Jean, and Johnny. This interconnected web of names shows the diverse ways in which names can correlate across different cultures and languages. \\
    \bottomrule
    \end{tabular}
    \caption{Easy and difficult Examples of MAGIC (1-conflict).}
    \label{tab:easy_hard_example}
\end{table*}

\end{document}